\documentclass[a4paper,5p,fleqn]{cas-dc}
\usepackage[numbers]{natbib}

\usepackage{amsmath,amssymb}
\usepackage{graphicx}
\usepackage{textcomp}
\usepackage{multirow}
\usepackage{hyperref}
\usepackage{float}
\usepackage{xcolor,soul,framed} 
\sethlcolor{yellow} 
\usepackage{pifont}

\usepackage{algpseudocode}

\usepackage{multicol}
\usepackage{subfig}
\usepackage{graphicx}

\usepackage[linesnumbered, ruled, vlined]{algorithm2e}
\usepackage{orcidlink}

\usepackage{nomencl} 
\makenomenclature
\renewcommand*\nompreamble{\begin{multicols}{2}}
\renewcommand*\nompostamble{\end{multicols}}

\def\BibTeX{{\rm B\kern-.05em{\sc i\kern-.025em b}\kern-.08em
    T\kern-.1667em\lower.7ex\hbox{E}\kern-.125emX}}
    
\usepackage{threeparttable}
\usepackage{mathtools}

\newcolumntype{P}[1]{>{\centering\arraybackslash}p{#1}}

\def\tsc#1{\csdef{#1}{\textsc{\lowercase{#1}}\xspace}}
\tsc{WGM}
\tsc{QE}
\tsc{EP}
\tsc{PMS}
\tsc{BEC}
\tsc{DE}

\begin{document}


\shorttitle{Edge-Enhanced Vision Transformer Framework for Accurate AI-Generated Image Detection}
\shortauthors{Das et~al.}

\title [mode=title] {Edge-Enhanced Vision Transformer Framework for Accurate AI-Generated Image Detection}                      

\credit{Conceptualization of this study, Methodology, Software}

\address{Department of Computer Science and Engineering, Uttara University, Dhaka, Bangladesh.}




\author{Dabbrata Das \orcidlink{0009-0008-0049-2048}}
\cormark[1]
\ead{dasdabbrata@gmail.com}

\author{Mahshar Yahan \orcidlink{0009-0001-0350-1666}}
\ead{yahanmahshar1@gmail.com}
\author{Md Tareq Zaman \orcidlink{0009-0006-8375-9734}}
\ead{tareqzaman777@gmail.com}
\author{Md Rishadul Bayesh \orcidlink{0009-0008-1955-5766}}
\ead{rishad.cse18.kuet@gmail.com}

\cortext[cor1]{Corresponding author}


\begin{abstract}
The rapid advancement of generative models has led to a growing prevalence of highly realistic AI-generated images, posing significant challenges for digital forensics and content authentication. Conventional detection methods mainly rely on deep learning models that extract global features, which often overlook subtle structural inconsistencies and demand substantial computational resources. To address these limitations, we propose a hybrid detection framework that combines a fine-tuned Vision Transformer (ViT) with a novel edge-based image processing module. The edge-based module computes variance from edge-difference maps generated before and after smoothing, exploiting the observation that AI-generated images typically exhibit smoother textures, weaker edges, and reduced noise compared to real images. When applied as a post-processing step on ViT predictions, this module enhances sensitivity to fine-grained structural cues while maintaining computational efficiency. Extensive experiments on the CIFAKE, Artistic, and Custom Curated datasets demonstrate that the proposed framework achieves superior detection performance across all benchmarks, attaining 97.75\% accuracy and a 97.77\% F1-score on CIFAKE, surpassing widely adopted state-of-the-art models. These results establish the proposed method as a lightweight, interpretable, and effective solution for both still images and video frames, making it highly suitable for real-world applications in automated content verification and digital forensics.
\end{abstract}

\maketitle





\section{Introduction}

The rapid advancement of generative models, particularly Generative Adversarial Networks (GANs)~\cite{gan} and diffusion-based architectures~\cite{dfa}, has revolutionized content creation by enabling the production of highly realistic AI-generated images. These developments have found extensive applications in creative industries, including digital art, gaming, entertainment, and content generation. However, the increasing realism of synthetic images poses serious challenges in digital forensics, cybersecurity, social media authenticity, and the detection of misinformation. AI-generated content can be exploited to manipulate public perception, spread false information, or facilitate fraud, making the ability to distinguish real images from AI-generated ones an urgent and critical problem.

Detecting AI-generated images remains challenging due to the high visual fidelity and subtle structural characteristics of modern synthetic content. Conventional deep learning-based classifiers, including convolutional neural networks (CNNs) and transformer-based models, primarily rely on global feature representations extracted from image patches or embeddings. While effective at capturing overall content patterns, these approaches often fail to identify fine-grained inconsistencies, such as minor edge misalignments, subtle texture artifacts, or slight deviations in structural regularity. Additionally, these models typically require large-scale training datasets and significant computational resources, which can limit their practicality in real-time or resource-constrained scenarios.

Prior research has explored a variety of detection strategies, ranging from CNN-based and transformer-based classifiers to statistical feature analysis and hybrid approaches combining multiple feature types. Despite promising performance under controlled conditions, these methods exhibit several limitations: limited sensitivity to subtle structural differences, dependency on large labeled datasets, and high computational complexity. Moreover, most approaches focus exclusively on image-level detection, leaving the extension to video content, where frames must be analyzed efficiently for temporal consistency, largely unexplored. Detecting AI-generated videos presents additional challenges, such as maintaining real-time inference speed, handling frame-to-frame variations, and scaling to high-resolution sequences, while still capturing the subtle cues that differentiate real from synthetic frames.

To address these challenges, there is a need for lightweight, interpretable, and efficient methods that can complement deep learning models. One promising direction (proposed) is the use of edge-based analysis, which exploits subtle structural discrepancies between real and AI-generated images. Typically, AI-generated images exhibit higher smoothness, reduced noise, and lower sharpness compared to real images, resulting in fewer and weaker edges. When standard image denoising or smoothing operations are applied, the edge structure of AI-generated images changes minimally, whereas real images, with naturally richer textures and sharper transitions, undergo more significant alterations. By quantifying these differences in edge behavior before and after smoothing, it is possible to derive a variance-based score that effectively highlights structural inconsistencies, providing a robust cue for distinguishing AI-generated content from real images. Such an approach can operate independently of complex deep learning models, reducing computational overhead while maintaining sensitivity to fine-grained features. When combined with deep learning-based global feature extraction, this hybrid strategy offers the potential to improve robustness, accuracy, and generalization across diverse datasets and image domains. Furthermore, because the edge-based approach is computationally inexpensive, it can be applied not only to static images but also to video frames, enabling near real-time AI video detection and analysis without significantly increasing inference time.

The proposed approach leverages this hybrid concept by integrating a fine-tuned Vision Transformer (ViT) with a novel edge-based image processing module. The ViT provides a global contextual understanding of image content, while the edge-based module focuses on local structural inconsistencies by computing edge-difference maps, variances, and derived scores. This combination allows the system to detect AI-generated content with high precision, even when global patterns appear nearly identical to real images. Beyond images, the method can be extended to video detection by processing individual frames efficiently, thereby maintaining fast inference speeds while ensuring temporal consistency across frames.

In summary, robust detection of AI-generated visual content requires methods that are sensitive to both global and local features, computationally efficient, and adaptable to multiple media types, including images and videos. Edge-based structural analysis, when combined with deep learning, provides a promising avenue to achieve these goals, enabling accurate, fast, and interpretable detection of AI-generated content across diverse scenarios.

The remainder of this paper is organized as follows. Section~\ref{sec:related_work} reviews the existing literature on AI-generated image detection, highlighting the limitations of current methods. Section~\ref{sec:methodology} describes the proposed hybrid framework in detail, including the fine-tuned ViT and the edge-based post-processing module. Section~\ref{sec:result} presents experimental results, ablation studies, and a comprehensive performance analysis. Finally, Section~\ref{sec:conclusion} concludes the paper and discusses potential directions for future research.


\begin{table}[!ht]
\caption{\textbf{Nomenclature:} Abbreviations and their definitions.}
\begin{framed}
\centering
\begin{tabular}{p{1.5cm}p{5.7cm}}
\textbf{Abbreviation} & \textbf{Definition} \\
\hline
AI & Artificial Intelligence \\
GAN & Generative Adversarial Network \\
CNN & Convolutional Neural Network \\
ViT & Vision Transformer \\
FFN & Feedforward Neural Network \\
MHSA & Multi-Head Self-Attention \\
ROC & Receiver-Operating Characteristic Curve \\
PRC & Precision-Recall Curve \\
EBP & Edge-Based Processing \\
CLIP & Contrastive Language-Image Pre-training \\
BLIP & Bootstrapped Language-Image Pretraining \\
AUC & Area Under the Curve \\
MAE & Masked Autoencoder \\
FLOPs & Floating Point Operations \\
VGG & Visual Geometry Group \\
IoU & Intersection over Union \\

\end{tabular}
\end{framed}
\end{table}

\section{Related Work}
\label{sec:related_work}

In recent years, research on detecting AI-generated images has grown rapidly. This is largely driven by the progress of generative models and the highly realistic images they can produce. In this section, we summarize the major approaches and findings from prior work.

\subsection{Traditional Statistical Methods}

The quality and variety of detection benchmarks have become crucial factors when evaluating AI-image detectors. Curated datasets such as GenImage~\cite{boychev2024bench}, Chameleon~\cite{yan2024sanity}, and ITW-SM~\cite{konstantinidou2025navigating} collect images from social media, various generative models (such as GANs, diffusion models, and transformers), and different domains to represent real-world situations. GenImage includes millions of samples from both open-source models and proprietary tools like Midjourney and Stable Diffusion, enabling comparisons across different generators. Despite their size, even the largest datasets struggle to keep up with the rapid development of new generative techniques, which quickly outdate existing benchmarks. The ITW-SM dataset~\cite{itwsm_dataset} focuses on challenges like distribution shifts and adversarial modifications, including images altered by social media compression and style transfer. In these cases, state-of-the-art models often see their accuracy drop by as much as 40\%, highlighting their vulnerability. Additionally, errors in labeling and differences between data domains lead to annotation bias in these datasets~\cite{jiang2023domainadapt}, making it difficult to compare results reliably across different studies. While people are working to improve data collection and continuous benchmarking, automating the labeling process is still a big challenge.

Watermarking methods~\cite{saberi2023robustness} add hidden patterns to images during creation or editing to prove where they came from. However, clever attackers can remove or hide these marks by cleaning, enhancing, or recreating the images using advanced techniques, making the watermarks ineffective. Detectors that rely on classifiers are also vulnerable to attacks like changing the image’s style, cropping, or rewording descriptions. Studies~\cite{boychev2024bench, zhang2024comprehensive} show that after such attacks, error rates for classifiers can double. Combining watermarking with multiple detection methods (looking at spatial, frequency, and semantic features) makes detection stronger but also makes the system more complex and less reliable in real-world situations. Currently, no single watermark or classifier can handle ongoing, smart attacks over long periods.

\subsection{Machine Learning Approaches}

Early AI-image detection mainly relied on convolutional neural networks (CNNs) to identify subtle statistical patterns unique to images generated by GANs. For example, Konstantinidou et al. \cite{konstantinidou2025navigating} present the DMID method which combines multiple types of features including color, texture, and deep features through multi-branch ensembles to better handle variations from different image generators. Their work highlights the importance of using diverse features to boost detector robustness. Liu et al. \cite{liu2023analysis} further explore this idea with multi-view fusion networks that integrate various feature types to enhance detection performance. Their analysis shows that combining different perspectives such as color and texture along with learned deep features helps improve resilience against different generator artifacts. However, they also point out that these CNN-based models often behave as “black boxes” making it difficult to understand how they make decisions and contributing to their weakness in adapting to new types of synthetic images. Gragnaniello et al. \cite{gragnaniello2021generalization} focus specifically on the generalization challenges in synthetic image detectors. They show that CNNs trained on one type of generative model struggle to detect images from new GANs or diffusion models. This gap reveals a key limitation because many CNN-based detectors fail when faced with distribution shifts or unseen image types which evolve rapidly as new generative architectures emerge. To overcome some of these limitations, newer methods like patch-based CNNs \cite{chen2024patchdet} and hybrid spectral-CNN ensembles \cite{karageorgiou2025spectral} analyze images at a finer level focusing on small patches or frequency components. These approaches improve detection especially with images that have been compressed or manipulated. However, such detailed analysis comes at the cost of longer processing times and heavier computational requirements making these methods less scalable for real-time applications.

\subsection{Deep Learning Approaches}

To address the diversity of modern generative models, recent strategies focus on patch-level and spectral learning. PatchDet~\cite{chen2024patchdet} segments images into overlapping patches and looks for repeated patterns or unusual artifacts. It is especially useful for detecting images made by diffusion and hybrid generators. Spectral learning~\cite{karageorgiou2025spectral} employs Fourier and Eigen decomposition to identify subtle differences that holistic classifiers might miss. These methods excel in identifying images with localized manipulations or hidden generator traces, showing up to 20\% improvement in mixed-domain detection. However, because they analyze many small regions, they can be slower and may struggle when images have been heavily edited as a whole. Ensemble patch-spectral architectures blend local and global features, but require careful calibration to avoid redundancy and computational overload.

With the increased use of diffusion-based generators, some detectors now focus on how well an image can be reconstructed from its compressed or hidden latent form. DIRE~\cite{wang2023dire} is one method that uses autoencoders to look for mistakes when trying to rebuild a synthetic image. This can reveal hidden smoothing or blending that is typical of generated images. LARE2~\cite{luo2024lare2} builds on this by using stepwise analysis of errors in the latent space. SeDID~\cite{ma2023sedid} uses a similar technique, analyzing errors at different steps in the image generation. These reconstruction-based methods work especially well on open-source diffusion and transformer generators. However, when faced with images from closed or proprietary models that behave differently during reconstruction, their accuracy can fall by as much as 30\%. Furthermore, these models need a lot of computational resources and often require careful tuning, which makes real-time use and scaling up difficult. 

Watermarking methods~\cite{saberi2023robustness} try to add hidden signals to images to mark their origin. However, attackers can often remove or hide these signals by cleaning the image, enhancing it or generating a new version. Detector systems can also be fooled with simple changes like cropping or switching the image style. For instance, practical evaluations~\cite{boychev2024bench} found that these attacks can easily make classifiers fail. Separate large-scale studies~\cite{zhang2024comprehensive} confirm that after such attacks, error rates can sometimes double. Mixing watermarking with multiple detection strategies does improve resilience to attacks, but this also makes the model more complex and harder to use in real-world situations. So far, no single watermark or classifier approach has proven stable against active, ongoing adversarial strategies.

To enable real-time and widespread use, there is growing research into simpler, lightweight deep learning models that can run on devices with less computing power. For example, Mittal et al.~\cite{mittal2024edgedeep} survey deep learning solutions built for medical imaging that work on the edge of images. Zhou et al.~\cite{zhou2021lightweight} demonstrate how compact CNNs can handle detection efficiently while saving memory. Albahri et al.~\cite{albahri2021lightweight} and Liu et al.~\cite{liu2023edge} show how pruning and quantization help shrink models for use on phones and smaller devices. Zou et al.~\cite{zou2023object} add that detection on edge devices is improving, but highlight that there’s often a trade-off. Lightweight models can come close to state-of-the-art accuracy for simple or clean images, but struggle with very subtle fakes or highly compressed samples. Advanced techniques like processing images at multiple scales and adjusting the inference pathway can help, but in general, robustness remains a challenge.

\subsection{Transformer-Based Approaches}

 Transformer-based models have greatly improved how we detect AI-generated images by understanding both the overall structure of an image and its meaning. For example, CLIP~\cite{radford2021learning} and DINO-V2~\cite{oquab2023dinov2} use large amounts of data to learn patterns between images and text, which helps them spot subtle signs that an image may be synthetic. Other models such as BLIP2~\cite{li2023blip}, ZeroVIL~\cite{liu2024zerovil}, and FAKEINVERSION~\cite{cazenavette2024fakeinversion} go further by combining image and language features, making it easier to handle new types of fake images or completely unseen data. These models use attention maps to highlight parts of an image that look unusual, such as strange textures, odd boundaries, or odd combinations of objects and text. This ability to use both visual and text clues also helps avoid the problem of models only working on the types of images they were trained on; for instance, BLIP2 and ZeroVIL perform well even with new generator styles~\cite{zhang2024comprehensive}.

Despite these advances, transformer models are still difficult to train and use in practice because they require a huge amount of data and computing power. They are also sensitive to attacks and messy datasets—problems like heavy image compression or changed captions can reduce their accuracy from 95\% down to 65\% in real-world tests~\cite{saberi2023robustness}.

To make these models even better, researchers have proposed new architectures. For instance, DeeCLIP~\cite{keetadeeclip2024} is a robust transformer model that maintains strong performance even when facing different types of generative sources. FatFormer~\cite{liuFatFormer2023} is another transformer specially designed for detecting fake images; it achieves high accuracy even when tested on new datasets. Stankovic et al.~\cite{StankovicViTInterior2025} show that transformers are also effective for specialized tasks like detecting AI-generated images in interior design. Sheng et al.~\cite{shengViLaCo2025} propose the ViLaCo model, which uses both visual and language clues in a weakly supervised way, so it does not need a lot of labeled data to find image forgeries. There are also hybrid transformer models that mix the standard Vision Transformer (ViT) with lighter, faster components like Linformer, allowing for accurate and efficient real-time detection on regular computers or devices~\cite{frontiersin2025}.

\subsection{Hybrid Approaches}

Recent research has shown clear advantages in combining different models through ensemble learning. For example, weighted ensembles merge outputs from multiple CNNs, transformers, and frequency-domain networks to boost overall detection accuracy~\cite{anan2025hybrid}. Similarly, Wolter et al.~\cite{wolter2023wavelet} demonstrate that stacking ensembles, which use meta-learners to combine predictions from base models, can significantly increase reliability in detecting AI-generated images. Gowda and Thillaiarasu~\cite{gowda2023ensemble} highlight the effectiveness of combining CNN architectures like SEResNet and XceptionNet with transformers in ensemble setups. Wodajo and Atnafu~\cite{wodajo2022cnn} add that frequency-domain features further strengthen these ensembles by exposing artifacts missed in other domains. Ricker et al.~\cite{ricker2024aeroblade} report that an ensemble called DFWild-Cup, which uses ResNet-34, DeiT, and XceptionNet combined with Haar wavelet features, achieves high performance with 93\% accuracy and 0.97 ROC-AUC~\cite{kafi2025hybrid}. However, ensembles typically require more computing resources and memory, which limits their use in real-time systems or on edge devices. Moreover, tuning ensemble weights involves extensive hyperparameter optimization, and these models may still struggle with new generator types unless specifically trained with adversarial techniques.

Self-supervised learning has recently become a powerful approach for improving model generalization across different generators. Zou et al.~\cite{zou2025bilevel} introduce BLADES, a bi-level optimization framework that customizes pretext tasks, such as EXIF tag prediction and manipulation detection, to enhance downstream AI-generated image detection. This method achieves state-of-the-art results, particularly in face detection, by combining EXIF-based linguistic features with surrogate detection tasks and dynamically adjusting task weights for maximum effectiveness. Zhou et al.~\cite{zhou2023masked} develop PATCHForensics using masked autoencoders (MAE), which improve a model’s ability to separate meaningful features from noisy data. Wolter et al.~\cite{wolter2023wavelet} apply frequency-based self-supervised learning techniques to further enhance feature discrimination. Nevertheless, self-supervised methods often depend on indirect cues. When the training objectives are not perfectly aligned with detection needs, their performance can fall behind fully supervised models, especially when detecting subtle manipulations or new types of synthetic images.

Overall, both ensemble learning and self-supervised learning represent important advances for AI-generated image detection, offering improved robustness and generalization when applied carefully.

Existing approaches to detecting AI-generated images face several challenges: datasets quickly become outdated and suffer from annotation bias; CNN-based and watermarking methods are vulnerable to adversarial modifications and fail to generalize across unseen generators; patch-wise and spectral methods improve detection but incur heavy computational costs; transformer-based models achieve strong accuracy but require large-scale pretraining and remain sensitive to perturbations; and hybrid ensembles, while robust, introduce complexity and high inference time. These limitations point to the need for solutions that are accurate, generalizable, interpretable, and lightweight for real-world deployment.

Motivated by these gaps, this paper introduces a hybrid framework that unifies ViT-based classification with edge difference analysis. The main contributions are:

\begin{itemize}
    \item Fine-tuning the base Vision Transformer on three different datasets, including a custom AI-generated image dataset, to enhance domain-specific representation.
    \item Proposing a novel edge-based image processing module that computes edge-difference maps and variance to classify real versus AI-generated images, offering interpretability and low computational overhead.
    \item Introducing the integration of the edge-based module as post-processing on ViT predictions, capturing structural inconsistencies overlooked by ViT alone.
    \item Demonstrating that the edge-based module alone is efficient and independent of deep learning, achieving strong performance with minimal computation.
    \item Presenting a robust hybrid framework for AI-generated image detection that achieves superior accuracy and practical applicability in real-world scenarios.
\end{itemize}



\begin{table*}[!htbp, align=\flushleft, width=\textwidth]
    \caption{Overview of significant contributions to AI-generated image detection}
    \centering
    \renewcommand{\arraystretch}{1.25}
    \footnotesize
    \begin{threeparttable}
    \begin{tabular}{P{2.6cm} P{2.4cm} P{2.6cm} P{3.4cm} P{2.0cm} P{2.5cm}}
        \hline \hline
        \textbf{Reference} & \textbf{Approach Type} & \textbf{Model / Framework} & \textbf{Technique / Contribution} & \textbf{Dataset} & \textbf{Performance / Limitation} \\
        \hline
        Boychev et al.~\cite{boychev2024bench} & Benchmarking & Resnet50, DeiT-S, Swin-T & Large-scale collection from GANs, diffusion, transformers, proprietary models & GenImage & Millions of samples; dataset quickly outdated; adversarial vulnerability \\
        \hline
        Konstantinidou et al.~\cite{konstantinidou2025navigating} & CNN-based & DMID (multi-branch CNN) & Combines color, texture, deep features via ensembles & ITW-SM & Better robustness across generators; still black-box, limited generalization \\
        \hline
        Liu et al.~\cite{liu2023analysis} & CNN-based & Multi-view fusion networks & Integrates color, texture, and deep features & Multiple & Improved resilience; weak adaptation to new generator artifacts \\
        \hline
        Gragnaniello et al.~\cite{gragnaniello2021generalization} & CNN-based & Cross-GAN Detection & Focus on generalization challenge & GAN/Diffusion datasets & Poor transferability to unseen generators \\
        \hline
        Chen et al.~\cite{chen2024patchdet} & Patch-based CNN & PatchDet & Patch-level detection of repeated patterns and artifacts & Diffusion & Strong performance, but computationally expensive \\
        \hline
        Karageorgiou et al.~\cite{karageorgiou2025spectral} & Spectral Learning & Hybrid CNN-Spectral & Fourier and Eigen decomposition for artifact localization & Mixed-domain & Up to 20\% improvement in cross-domain detection; slower processing \\
        \hline
        Wang et al.~\cite{wang2023dire} & Reconstruction-based & DIRE (Autoencoder) & Reconstruction error detection in latent space & Diffusion & Effective on open-source models; 30\% drop on proprietary models \\
        \hline
        Luo et al.~\cite{luo2024lare2} & Reconstruction-based & LARE2 & Stepwise latent error analysis & Diffusion & Fine detection; high computational needs \\
        \hline
        Mittal et al.~\cite{mittal2024edgedeep}, Zhou et al.~\cite{zhou2021lightweight} & Lightweight CNNs & Edge-optimized CNNs & Pruning, quantization, efficient architectures & Medical, Edge datasets & Near SOTA on clean data; weaker on subtle or compressed fakes \\
        \hline
        Radford et al.~\cite{radford2021learning}, Oquab et al.~\cite{oquab2023dinov2} & Transformer-based & CLIP, DINO-V2 & Large-scale pretraining with text-image pairs & Web-scale & Robust zero-shot, but expensive; vulnerable to perturbations \\
        \hline
        Li et al.~\cite{li2023blip}, Liu et al.~\cite{liu2024zerovil} & Vision-Language Transformers & BLIP2, ZeroVIL & Multimodal alignment for cross-domain detection & Multiple & Strong domain transfer; high pretraining cost \\
        \hline
        Cazenavette et al.~\cite{cazenavette2024fakeinversion}, Keetadee et al.~\cite{keetadeeclip2024} & Robust Transformers & FAKEINVERSION, DeeCLIP & Robust cross-generator detection & Benchmarks & Accuracy drop under compression; DeeCLIP improves stability \\
        \hline
        Liu et al.~\cite{liuFatFormer2023} & Transformer-based & FatFormer & Adaptive transformer tailored for detection & Cross-domain datasets & High generalization; computationally heavy \\
        \hline
        Zou et al.~\cite{zou2025bilevel} & Self-supervised & BLADES & Bi-level optimization, EXIF-based pretexts & Faces & SOTA on face datasets; depends on pretext choice \\
        \hline
        Wolter et al.~\cite{wolter2023wavelet}, Kafi et al.~\cite{kafi2025hybrid} & Hybrid Ensembles & CNN + Transformer + Wavelet & Weighted/stacking ensembles with meta-learners & Mixed datasets & High accuracy (up to 93\%, AUC 0.97); increased complexity and overhead \\
        \hline
    \end{tabular}
    \end{threeparttable}
    \label{tab:ai_image_detection}
\end{table*}

\section{Methodology}
\label{sec:methodology}
This section presents a detailed description of the data preprocessing steps and the overall model architecture. It also introduces a newly proposed module designed to enhance model robustness through fine-tuning of the Vision Transformer (ViT).

\subsection{Data Preprocessing}
All input images are converted to the RGB color space and resized to $224 \times 224$ pixels, the standard input resolution for Vision Transformer (ViT) models, using Lanczos resampling to preserve quality. The images are then normalized and transformed into tensor format via the AutoImageProcessor from the Hugging Face Transformers library. Error handling is incorporated to skip unreadable or corrupted images, ensuring dataset integrity before training.

\subsection{Model Architecture}
\begin{figure}[htbp]
    \centering
    \includegraphics[width=0.9\linewidth]{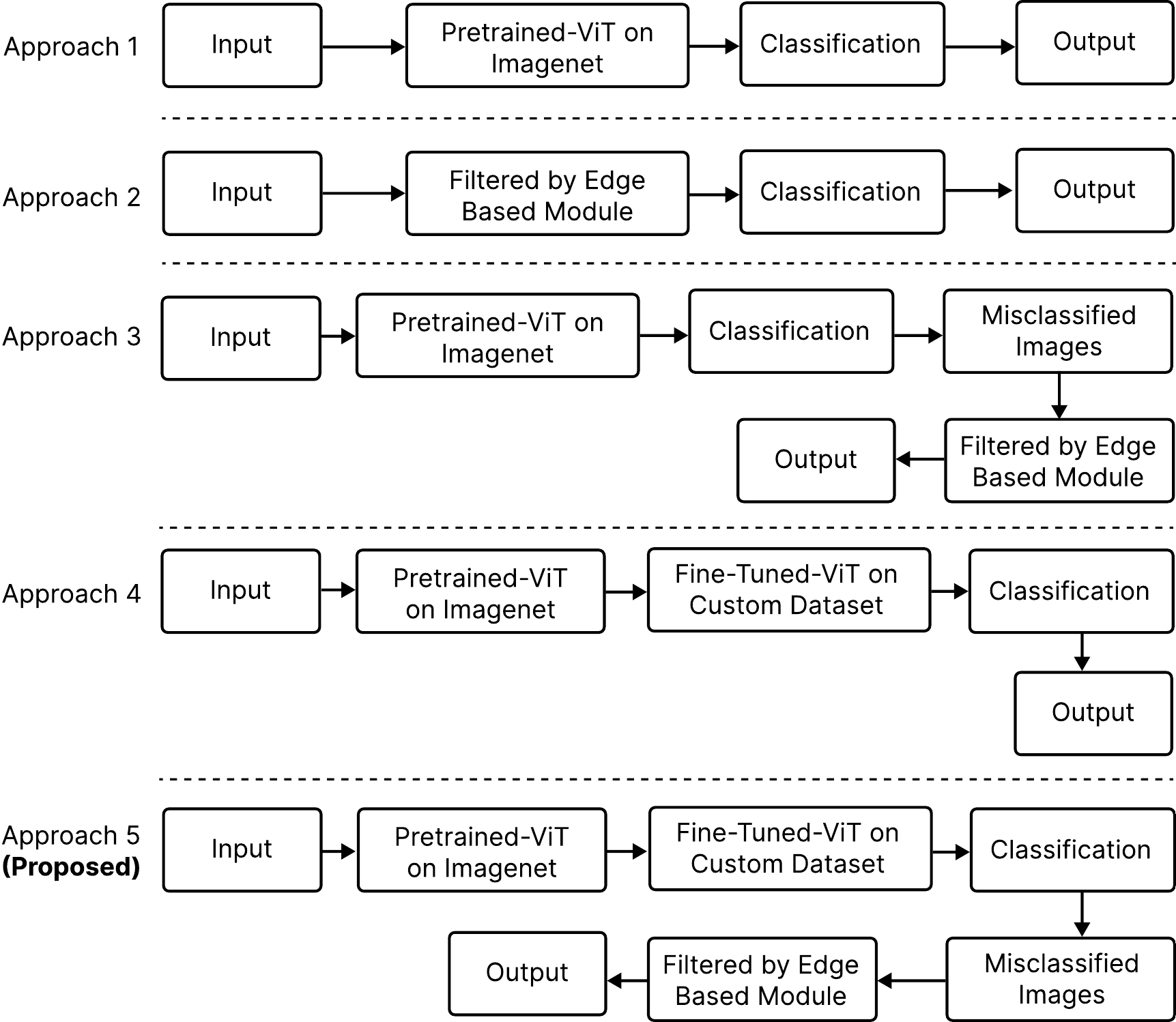}
    \caption{All implemented pipelines: (1) baseline ViT, (2) standalone edge-based module, (3) pretrained ViT with edge features, (4) fine-tuned ViT, and (5) the proposed fine-tuned ViT with edge-based post-processing.}
    \label{fig:all_methods}
\end{figure}

To implement the proposed work, we employed five different approaches: using only a Vision Transformer (ViT) model~\cite{vit_base} pretrained on the ImageNet dataset~\cite{imagenet_dataset}, using only the proposed edge-based processing module, applying the edge-based processing module as a post-processing step to the pretrained ViT, using a fine-tuned ViT, and combining the fine-tuned ViT with the edge-based processing module. Figure~\ref{fig:all_methods} illustrates all the approaches we implemented, including the proposed method.

\subsubsection{Vision Transformer (ViT) Model}
In this study, we employ the Vision Transformer (ViT) base~\cite{vit_base} architecture pre-trained on the large-scale ImageNet-21k dataset~\cite{imagenet_dataset}. The ViT model processes an input image by dividing it into non-overlapping patches of size $16 \times 16$ pixels, which are then flattened and linearly projected into a sequence of patch embeddings. These embeddings are augmented with learnable position embeddings and passed through a standard Transformer encoder.

Given an image $x \in \mathbb{R}^{H \times W \times C}$, it is first reshaped into $N$ patches $x_p \in \mathbb{R}^{P^2 \cdot C}$, where $H$ and $W$ are the image height and width, $C$ is the number of channels, $P$ is the patch size, and $N = \frac{HW}{P^2}$ is the number of patches. The patch embedding is computed as:
\begin{equation}
z_0 = [x_{\text{cls}}; x_p^1E; x_p^2E; \dots; x_p^N E] + E_{\text{pos}}
\end{equation}
where $x_{\text{cls}}$ is a learnable classification token, $E$ is the patch embedding projection, and $E_{\text{pos}}$ represents position embeddings. The sequence is processed through $L$ Transformer encoder layers, each consisting of Multi-Head Self-Attention (MHSA) and a feed-forward network (FFN):
\begin{equation}
\text{MHSA}(Q, K, V) = \text{softmax}\left(\frac{QK^\top}{\sqrt{d_k}}\right)V
\end{equation}
The output corresponding to the classification token is fed into a classification head to produce the final prediction.

Initially, we evaluate the pre-trained model directly to measure baseline performance using metrics such as accuracy, precision, recall, and F1-score. We then fine-tune the model on three different datasets, including our own customized dataset, to enhance its ability to distinguish AI-generated images from real images. This fine-tuning process allows the model to adapt its learned representations to the domain-specific characteristics of the target task, improving classification robustness.

\subsubsection{Edge-Based Processing Module}
\begin{figure}[htbp]
    \centering
    \includegraphics[width=0.9\linewidth]{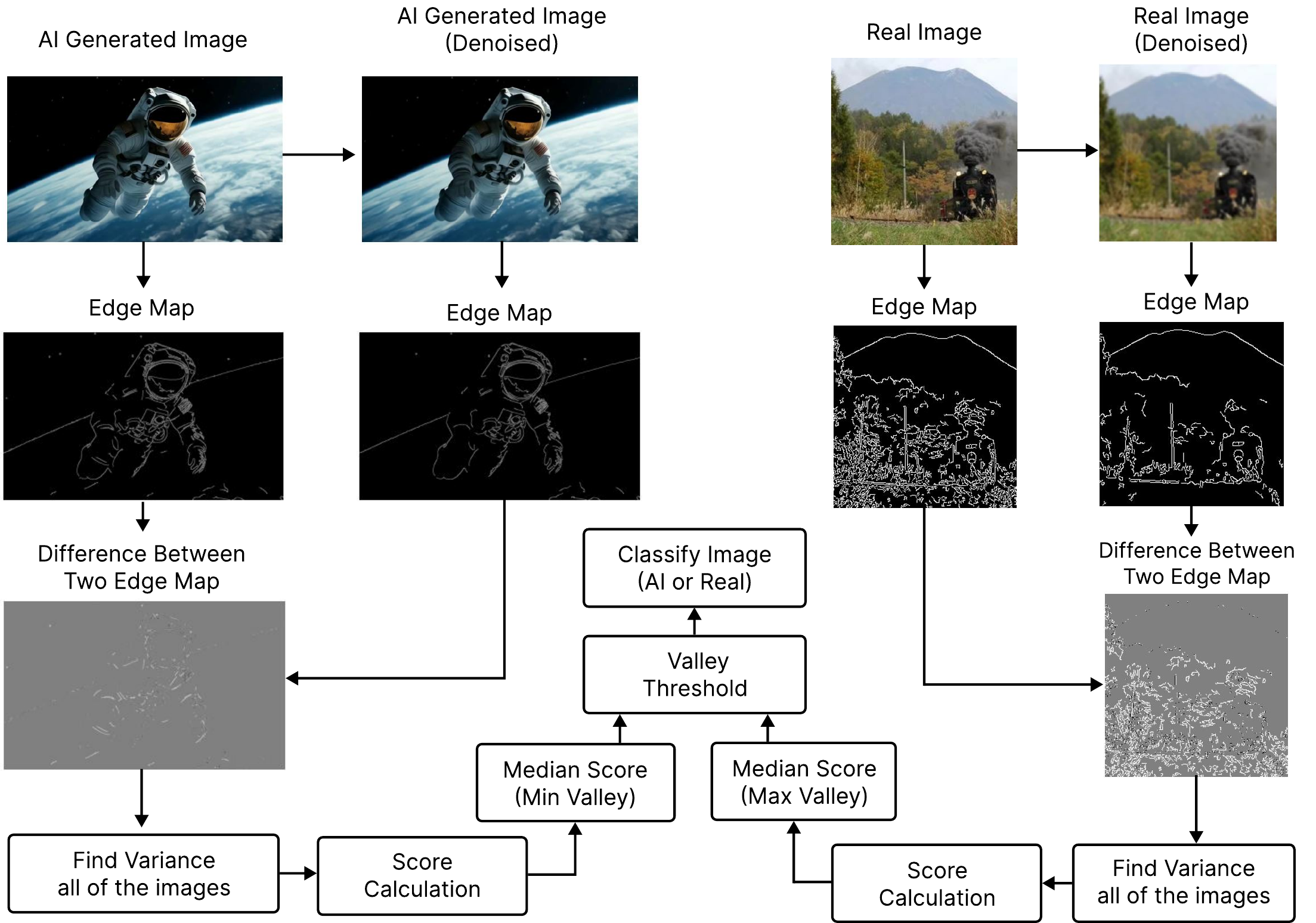}
    \caption{Proposed edge-based module to classify images as AI-generated or real.}
    \label{fig:eb_module}
\end{figure}

The proposed Edge-Based Processing Module (Figure~\ref{fig:eb_module}) aims to capture subtle differences in edge structures between real and AI-generated images. This module operates in three stages: edge detection, score calculation, and threshold determination.

\textbf{Edge Detection.}
Given an input RGB image $I(x, y)$, we first convert it to grayscale:
\begin{equation}
    I_g(x, y) = 0.299 R + 0.587 G + 0.114 B
\end{equation}

To reduce high-frequency noise, we apply Gaussian smoothing:
\begin{equation}
I_b(x, y) = I_g(x, y) * G_{\sigma}
\end{equation}
where $G_{\sigma}$ is a Gaussian kernel of size $3 \times 3$ with standard deviation $\sigma$.

Edges are extracted using the Canny edge detector applied to both the original grayscale image $I_g$ and the blurred image $I_b$:
\begin{equation}
E_1 = \text{Canny}(I_g, T_{low}, T_{high})
\end{equation}
\begin{equation}
E_2 = \text{Canny}(I_b, T_{low}, T_{high})
\end{equation}
where $T_{low}$ and $T_{high}$ are the lower and upper thresholds for edge detection.

\textbf{Score Calculation.}
We compute the difference map between the two edge images:
\begin{equation}
D = E_1 - E_2
\end{equation}
The mean of the difference map is:
\begin{equation}
\mu_D = \frac{1}{N} \sum_{i=1}^{N} D_i
\end{equation}
and the variance is:
\begin{equation}
\sigma_D^2 = \frac{1}{N} \sum_{i=1}^{N} (D_i - \mu_D)^2
\end{equation}
Let $N_{\text{edges}}$ denote the total number of edge pixels in $E_1$. The \textit{edge variance score} $S$ is defined as:
\begin{equation}
S = \frac{N_{\text{edges}}}{\sigma_D^2 + \epsilon}
\end{equation}
where $\epsilon$ is a small constant to avoid division by zero. Here, the variance term is placed in the denominator to normalize the measure with respect to the number of non-zero pixels. This normalization is necessary because images containing only a few objects naturally exhibit low variance; dividing by the variance yields a ratio that remains comparable across images and datasets. Without this division, establishing a consistent threshold for different datasets would not be feasible.

\textbf{Threshold Determination.}
To separate real and AI-generated images, scores are computed for a calibration set of both classes. Let $\tilde{S}_r$ and $\tilde{S}_f$ be the median scores of real and fake images, respectively. We build a histogram of all scores and define the \textit{valley region} between $\min(\tilde{S}_r, \tilde{S}_f)$ and $\max(\tilde{S}_r, \tilde{S}_f)$. The optimal decision threshold $T$ is selected as:
\begin{equation}
T = \underset{s \in \text{Valley Region}}{\arg\min} \; H(s)
\end{equation}
where $H(s)$ denotes the histogram bin count at score $s$. An image is classified as \textit{fake} if $S \geq T$ and \textit{real} otherwise.

\begin{algorithm}[h]
\small
\caption{Optimal Valley Threshold Computation}
\SetAlgoLined
\SetKwInOut{Input}{Input}
\SetKwInOut{Output}{Output}

\Input{Real scores $S_R$, Fake scores $S_F$}
\Output{Threshold $T$}

\If{$|S_R| = 0$ \textbf{or} $|S_F| = 0$}{\Return None}
$\text{median}_R \gets \text{median}(S_R)$\;
$\text{median}_F \gets \text{median}(S_F)$\;
$S \gets S_R \cup S_F$\;
$(\text{counts}, \text{bin\_edges}) \gets \text{histogram}(S, 50)$\;
$\text{bin\_centers} \gets (\text{bin\_edges}[:-1] + \text{bin\_edges}[1:])/2$\;
$\text{valley\_region} \gets$ counts between $\text{median}_R$ and $\text{median}_F$\;
$\text{valley\_bins} \gets$ corresponding bin centers\;
\If{$|\text{valley\_region}| = 0$}{\Return $(\text{median}_R + \text{median}_F) / 2$}
$T \gets \text{valley\_bins}[\arg\min(\text{valley\_region})]$\;

\Return $T$\;

\end{algorithm}

\begin{figure*}[htbp]
    \centering
    \includegraphics[width=0.98\linewidth]{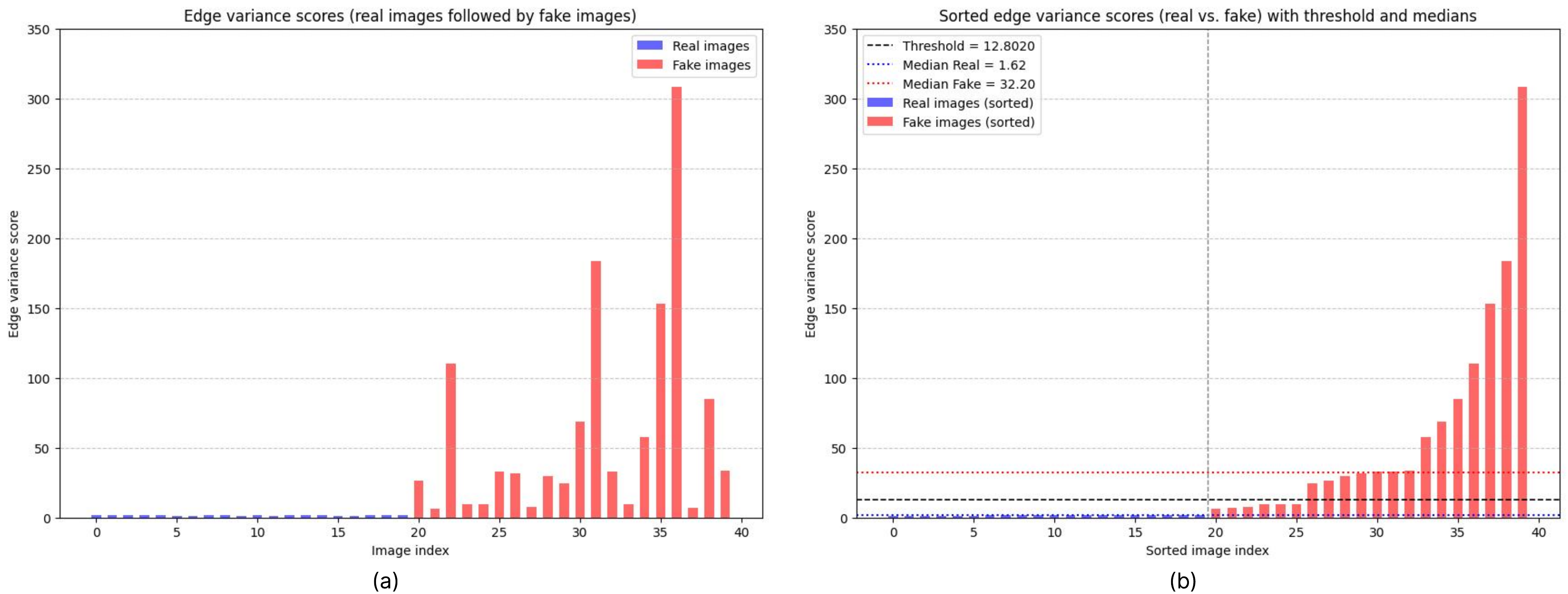}
    \caption{Visualization of edge variance scores for real and fake samples: (a) unsorted distributions, (b) sorted scores with medians, where the threshold is chosen between the two medians using the valley-point strategy.}
    \label{fig:find_th}
\end{figure*}








\IncMargin{1em}
\begin{algorithm}[h]
\small
\caption{Edge-Based Image Classification for Real vs AI-Generated Images}
\label{alg:edge_based_classification_with_threshold}
\SetAlgoLined
\SetKwInOut{Input}{Input}
\SetKwInOut{Output}{Output}

\Input{Image set $\mathcal{I}$, decision threshold $T$ (optional; computed if not provided)}
\Output{Predicted labels for $\mathcal{I}$}

\BlankLine
\textbf{Step 1: Compute edge variance scores} \\
\ForEach{image $I$ in $\mathcal{I}$}{
    Convert to grayscale and apply Gaussian blur\;
    Compute Canny edges on original and blurred images\;
    Compute difference $D$, variance of $D$, and number of edge pixels $N$\;
    $S_I \gets N / (\text{variance} + \epsilon)$\;
}

\textbf{Step 2: Determine classification threshold (if not given)} \\
\If{$T$ not provided}{
    Separate scores into $S_R$ (real) and $S_F$ (fake) for calibration\;
    \If{$S_R$ or $S_F$ is empty}{\Return None}
    Compute medians $\text{median}_R$, $\text{median}_F$\;
    Compute histogram of all scores and select valley between medians\;
    \If{valley region is empty}{\Return $(\text{median}_R + \text{median}_F)/2$}
    $T \gets$ bin corresponding to minimum in valley region\;
}

\textbf{Step 3: Classification} \\
\ForEach{$I \in \mathcal{I}$}{
    label $\gets$ Real if $S_I < T$, else AI-generated
}

\Return predicted labels\;

\end{algorithm}

\subsubsection{Fine-Tuned ViT with Edge-Based Processing}

In the proposed approach, we integrate fine-tuning of the Vision Transformer (ViT) with the Edge-Based Processing Module to enhance the classification of AI-generated and real images. We employ the \texttt{vit-base-patch16-224-in21k} architecture~\cite{vit_base}, initially pre-trained on the large-scale ImageNet-21k dataset~\cite{imagenet_dataset}, and fine-tune it on three distinct datasets, including our customized dataset specifically curated for AI image detection. This fine-tuning process allows the model to adapt its learned representations to the unique texture, color distribution, and spatial characteristics of the target domain, thereby improving its discriminative capability.

After fine-tuning, the model generates classification predictions for the test set using standard evaluation metrics such as accuracy, precision, recall, and F1-score. While the fine-tuned ViT achieves high performance, certain challenging samples, particularly those with minimal texture discrepancies between real and AI-generated content, remain misclassified. To address this limitation, we introduce a post-processing refinement step using the proposed Edge-Based Processing Module.

In this refinement stage, the misclassified images from the fine-tuned ViT are passed through the edge-based analysis pipeline, which consists of edge detection, score calculation, and threshold-based classification. Specifically, the module extracts structural edge patterns from both the original and Gaussian-smoothed grayscale versions of the image, computes the edge variance score, and applies a decision threshold determined from a calibration set. This process effectively captures subtle structural inconsistencies often overlooked by the ViT's global patch-based representation.

\begin{figure*}[htbp]
    \centering
    \includegraphics[width=0.90\linewidth]{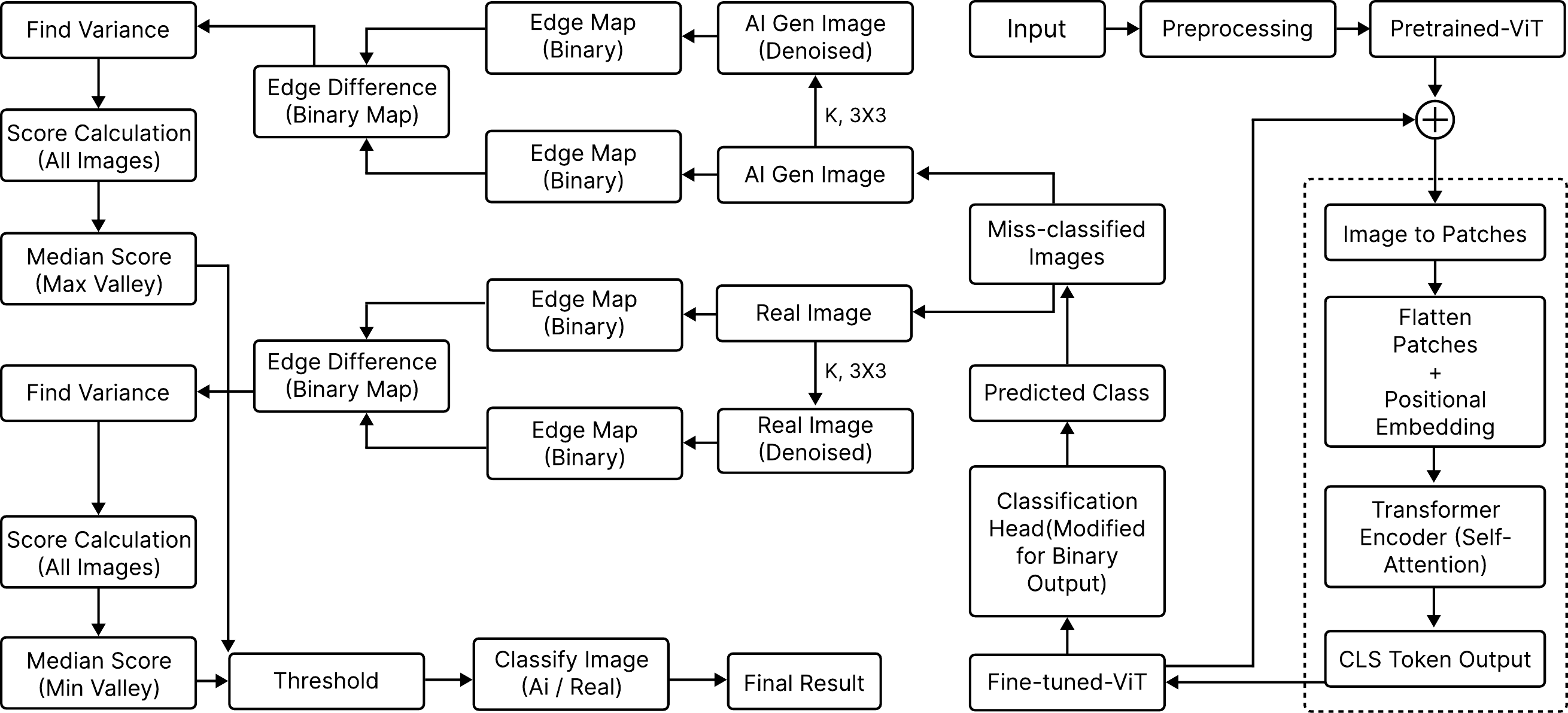}
    \caption{Overall pipeline of the proposed method, where the fine-tuned ViT is combined with edge-based post-processing.}
    \label{fig:whole_pipeline}
\end{figure*}

By reclassifying only the misclassified samples using the secondary edge-based module, the proposed approach integrates the ViT’s capacity to capture global contextual patterns with the edge-based module’s sensitivity to subtle structural variations, thereby improving classification performance and enhancing overall robustness. Figure \ref{fig:whole_pipeline} depicts the pipeline of the proposed method.

\IncMargin{1em}
\begin{algorithm}[h]
\small
\caption{Two-Stage Classification: Fine-Tuned ViT with Edge-Based Refinement}
\label{alg:vit_edge_pipeline}
\SetAlgoLined
\SetKwInOut{Input}{Input}
\SetKwInOut{Output}{Output}

\Input{Image set $\mathcal{I}$, fine-tuned ViT model $M_{\text{ViT}}$, edge-based module $M_{\text{Edge}}$}
\Output{Predicted labels for $\mathcal{I}$}

\BlankLine
\textbf{Stage 1: ViT Classification} \\
\ForEach{image $I$ in $\mathcal{I}$}{
    Predict label $y_{\text{ViT}} \gets M_{\text{ViT}}(I)$\;
    Store $y_{\text{ViT}}$ in prediction set $\mathcal{P}$\;
}

\textbf{Identify Misclassified Samples (Calibration Phase)} \\
Compare $\mathcal{P}$ with ground truth to find misclassified set $\mathcal{M}$\;

\textbf{Stage 2: Edge-Based Refinement} \\
\ForEach{image $I$ in $\mathcal{M}$}{
    Convert $I$ to grayscale and apply Gaussian blur\;
    Perform Canny edge detection on original and blurred images\;
    Compute edge variance score $S_I$\;
    \If{$S_I < T$}{label $\gets$ Real} \Else{label $\gets$ AI-generated}
    Update prediction in $\mathcal{P}$\;
}

\Return $\mathcal{P}$\;

\end{algorithm}

\section{Implementation}
\label{sec:implementation}
\subsection{Datasets}
We conducted our experiments using three balanced image datasets comprising both AI-generated and real images. The first dataset~\cite{artistic_dataset} consists of AI-generated images created by state-of-the-art image synthesis models such as DALL-E and Midjourney, along with real images produced by humans. The AI-generated portion predominantly contains artistic and non-photorealistic content, as an imbalanced inclusion of artistic works in the real-image category was found to improve classification accuracy. However, we observed that the presence of noise elements, such as film grain or fur textures, increased misclassification rates, regardless of the image origin. This phenomenon is likely due to the intrinsic noise-based generation process of diffusion models, which can cause the classifier to overfit to noise patterns. The subset used in our work included 3,780 real and 4,000 AI-generated samples, totaling 7,780 images.

The second dataset~\cite{cifake_dataset} is derived from a benchmark collection of real images from the CIFAR-10~\cite{cifar10_dataset} dataset and their synthetic counterparts generated with Stable Diffusion v1.4~\cite{stable_diffusion}. From the original large-scale version, we selected a balanced subset containing 4,000 real and 4,000 synthetic images, maintaining the original resolution of $32 \times 32 \times 3$ to preserve fine-grained texture details essential for detection tasks.

The third dataset is a custom-curated collection containing 2500 real and 2500 AI-generated images, representing a wide range of visual domains, including highly realistic photographs and stylized synthetic artworks. This dataset was specifically designed to assess the model’s cross-domain generalization capability.

All datasets underwent a uniform preprocessing procedure that included pixel-value normalization to the range $[0,1]$, horizontal flipping for data augmentation during training, and aspect ratio preservation where applicable. The combination of diverse artistic and photographic content, varying resolutions, and differing noise characteristics provided a robust foundation for evaluating the proposed model’s performance and resilience across multiple image domains.

\begin{figure}[htbp]
    \centering
    \includegraphics[width=0.9\linewidth]{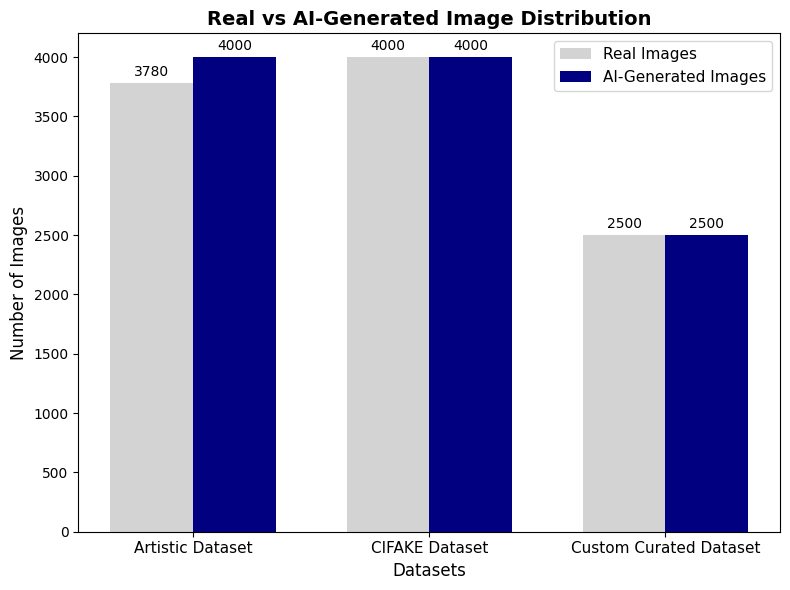}
    \caption{Comparison of the number of real and AI-generated images across three datasets: Artistic Dataset~\cite{artistic_dataset}, CIFAKE Dataset~\cite{cifake_dataset}, and a Custom Curated Dataset.}
    \label{fig:dataset}
\end{figure}

\subsection{Environmental Setup}

All experiments were conducted using PyTorch~\cite{pytorch} on a dual NVIDIA T4 GPU setup, with each GPU having 16 GB of VRAM. The proposed approach was applied to a base Vision Transformer (ViT) model for fine-tuning on three datasets: the Artistic dataset~\cite{artistic_dataset}, CIFAKE~\cite{cifake_dataset}, and a custom curated dataset.
Before fine-tuning, the pretrained ViT model was evaluated using a held-out test set, where performance metrics such as accuracy, precision, recall, and F1-score were computed.
For fine-tuning, the Hugging Face Trainer API was utilized with the following hyperparameters: a learning rate of $1\times 10^{-5}$, weight decay of $0.05$, batch size of $32$ for both training and evaluation, and $5$ training epochs. FP16 mixed-precision training was enabled to optimize GPU memory usage and computation speed. The F1-score was used as the primary metric for selecting the best model.
The average training durations were approximately $1$ hour for the Artistic dataset, $0.5$ hours for CIFAKE, and $0.7$ hours for the custom dataset. Throughout all experiments, the same hardware configuration was maintained to ensure a fair comparison.

\subsection{Evaluation metrics}

The evaluation process consisted of two stages: baseline evaluation and post-processing evaluation. 
In the baseline stage, the pretrained or fine-tuned model was tested on the held-out test set, where predictions were obtained for each sample. 
The predicted labels were compared against the ground-truth labels to compute the standard classification metrics.

The primary metrics used in our evaluation were defined as follows:

\begin{itemize}
    \item \textbf{Accuracy:} The proportion of correctly classified samples:
    \begin{equation}
        \text{Accuracy} = \frac{TP + TN}{TP + TN + FP + FN}
    \end{equation}
    
    \item \textbf{Precision:} The proportion of correctly predicted positive samples among all predicted positives:
    \begin{equation}
        \text{Precision} = \frac{TP}{TP + FP}
    \end{equation}
    
    \item \textbf{Recall:} The proportion of correctly predicted positive samples among all actual positives:
    \begin{equation}
        \text{Recall} = \frac{TP}{TP + FN}
    \end{equation}
    
    \item \textbf{F1-score:} The harmonic mean of precision and recall:
    \begin{equation}
        \text{F1-score} = \frac{2 \times \text{Precision} \times \text{Recall}}{\text{Precision} + \text{Recall}}
    \end{equation}

    \item \textbf{Intersection over Union (IoU):} For each class, the IoU is given by:
    \begin{equation}
        \text{IoU}_{c} = \frac{TP_{c}}{TP_{c} + FP_{c} + FN_{c}}
    \end{equation}
    where $TP_{c}$, $FP_{c}$, and $FN_{c}$ are computed for class $c$.

    \item \textbf{Mean IoU (mIoU):} The average IoU across all $C$ classes:
    \begin{equation}
        \text{mIoU} = \frac{1}{C} \sum_{c=1}^{C} \text{IoU}_{c}
    \end{equation}
\end{itemize}

In addition to baseline evaluation, a post-processing step based on edge variance analysis was applied to misclassified samples. 
An optimal threshold, calibrated from real and fake training images, was used to re-evaluate and potentially correct these samples. 
Performance gains were measured by: (i) corrected misclassifications, (ii) changes in accuracy, F1-score, and mIoU, and (iii) percentage accuracy improvement:
\begin{equation}
    \text{Accuracy Improvement} = \frac{\text{Acc}_{\text{post}} - \text{Acc}_{\text{baseline}}}{\text{Acc}_{\text{baseline}}} \times 100
\end{equation}

This two-stage evaluation quantified both the baseline performance and the improvements obtained through targeted error correction.

\section{Results}
\label{sec:result}

\subsection{Comparison with State-of-the-Art Architectures}

\subsubsection{Quantitative Analysis}

To assess the effectiveness of the proposed module, we compare it against widely adopted architectures, including ResNet50~\cite{resnet50}, MobileNetV2~\cite{mobilenetv2}, DenseNet121~\cite{densenet121}, EfficientNet-B0~\cite{efficientnetb0}, VGG19~\cite{vgg19}, AlexNet~\cite{alexnet}, GoogLeNet~\cite{googlenet}, and ViT~\cite{vit_base}. Table~\ref{tab:sota_comparison} summarizes the results in terms of model complexity, computational cost, inference time, and classification metrics. Notably, the proposed ViT + EBP achieves the highest F1-score and accuracy, while maintaining competitive precision and recall.

\begin{figure}[htbp]
    \centering
    \includegraphics[width=1.0\linewidth]{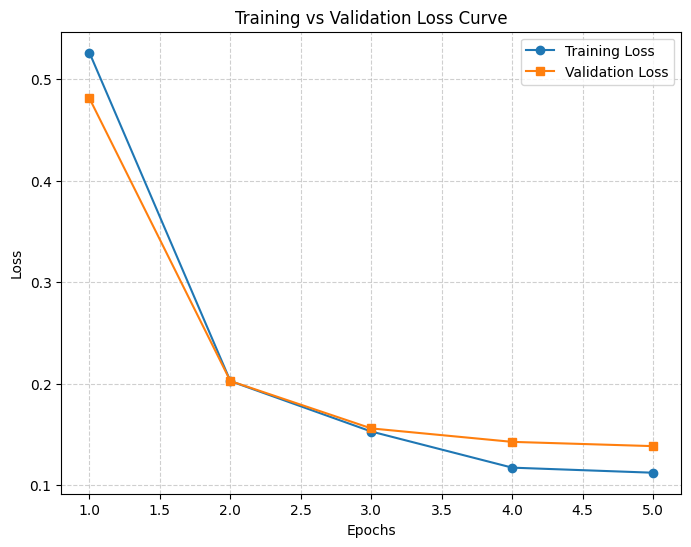}
    \caption{Training and validation loss curves during ViT fine-tuning on the CIFAKE~\cite{cifake_dataset} dataset, showing smooth convergence and minimal overfitting.}
    \label{fig:loss_curve}
\end{figure}

Furthermore, the training and validation loss curves, illustrated in Figure~\ref{fig:loss_curve}, demonstrate smooth and stable convergence, with minimal gap between training and validation performance. This indicates that the proposed method not only achieves superior classification results compared to baseline models but also generalizes effectively without overfitting.

\begin{table*}[htbp]
\centering
\caption{Performance evaluation of the proposed method compared with state-of-the-art architectures on the CIFAKE dataset~\cite{cifake_dataset} in terms of model size, FLOPs, inference time, and classification metrics.}
\label{tab:sota_comparison}
\resizebox{\textwidth}{!}{
\begin{tabular}{lccccccccc}
\hline
\textbf{Method} & \textbf{Params} & \textbf{FLOPs} & \textbf{Inference Time (s)} & \textbf{Precision} & \textbf{Recall} & \textbf{F1-Score} & \textbf{Accuracy} \\
\hline
ResNet50~\cite{resnet50}         & 23.51M  & 4.13 GMac   & 0.0047 & 0.9650 & 0.8275 & 0.8910 & 0.8988 \\
MobileNetV2~\cite{mobilenetv2}      & 2.23M   & 0.32 GMac    & 0.0028 & 0.9439 & 0.9227 & 0.9332 & 0.9337 \\
DenseNet121~\cite{densenet121}      & 6.96M   & 2.90 GMac   & 0.0050 & 0.9123 & 0.9698 & 0.9402 & 0.9387 \\
EfficientNet-B0~\cite{efficientnetb0}  & 4.01M   & 0.41 GMac & 0.0033 & 0.9385 & 0.9150 & 0.9266 & 0.9275 \\
VGG19~\cite{vgg19}            & 139.58M & 19.68 GMac  & 0.0082 & 0.9317 & 0.9455 & 0.9386 & 0.9375 \\
AlexNet~\cite{alexnet}          & 57.01M  & 0.712 GMac & 0.0024 & 0.9676 & 0.8732 & 0.9179 & 0.9200 \\
GoogLeNet~\cite{googlenet}        & 5.60M   & 1.51 GMac   & 0.0030 & 0.9147 & 0.9279 & 0.9212 & 0.9175 \\
\textbf{ViT + EBP (Proposed)} & 86M & 8.69 GMac & 0.0919 & \textbf{0.9689} & \textbf{0.9867} & \textbf{0.9777} & \textbf{0.9775} \\
\hline
\end{tabular}}
\end{table*}

\subsubsection{Qualitative Analysis}
In addition to reporting quantitative results, we conduct a qualitative analysis to assess model efficiency, classification reliability, and interpretability. A summary of this comparison is provided in Table~\ref{tab:qualitative_comparison}, where a simple checkmark and cross representation is used. The proposed ViT + EBP framework, though not the most lightweight due to its relatively higher parameter count, consistently delivers superior accuracy, robustness across datasets, and clear visual separability between real and AI-generated images. In contrast, lightweight architectures such as MobileNetV2~\cite{mobilenetv2} and EfficientNet-B0~\cite{efficientnetb0} are efficient in terms of FLOPs and inference time but sacrifice robustness and visual interpretability. Similarly, classical CNNs such as VGG19~\cite{vgg19} and AlexNet~\cite{alexnet} achieve moderate classification accuracy but fall short in terms of model efficiency and separability. These observations highlight the importance of integrating edge-based processing into transformer architectures, which not only improves numerical performance but also enhances qualitative distinctions essential for explainability in real-world scenarios.  

The proposed edge-based processing approach builds upon an edge variance score, derived from the variance between two edge maps: one generated before denoising and the other after denoising. Fake images typically yield lower variance scores, as their edge structures remain relatively unchanged across denoising. In contrast, real images exhibit higher variance due to noticeable differences in edge structures. Applying this procedure across all samples produces edge difference maps and corresponding variance scores, which enable reliable discrimination between real and AI-generated content.  

\begin{figure*}[htbp]
    \centering
    \includegraphics[width=1.0\linewidth]{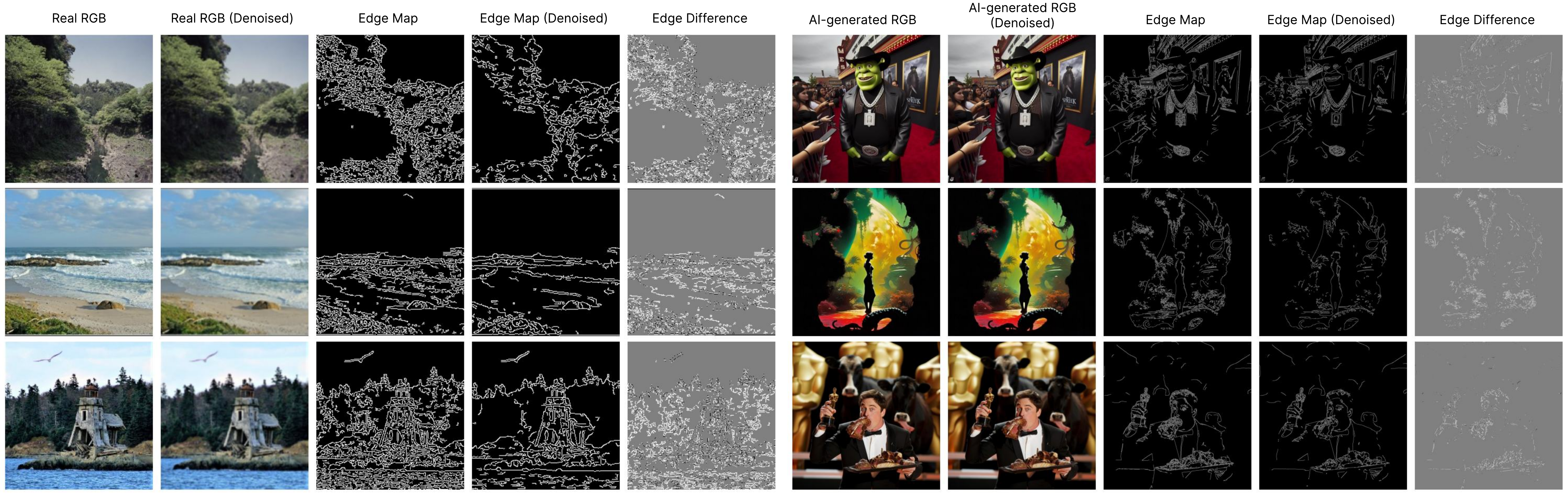}
    \caption{Examples of real and AI-generated images with their denoised versions, corresponding edge maps, and variance maps.}
    \label{fig:ed_maps}
\end{figure*}

\begin{figure*}[htbp]
    \centering
    \includegraphics[width=1.0\linewidth]{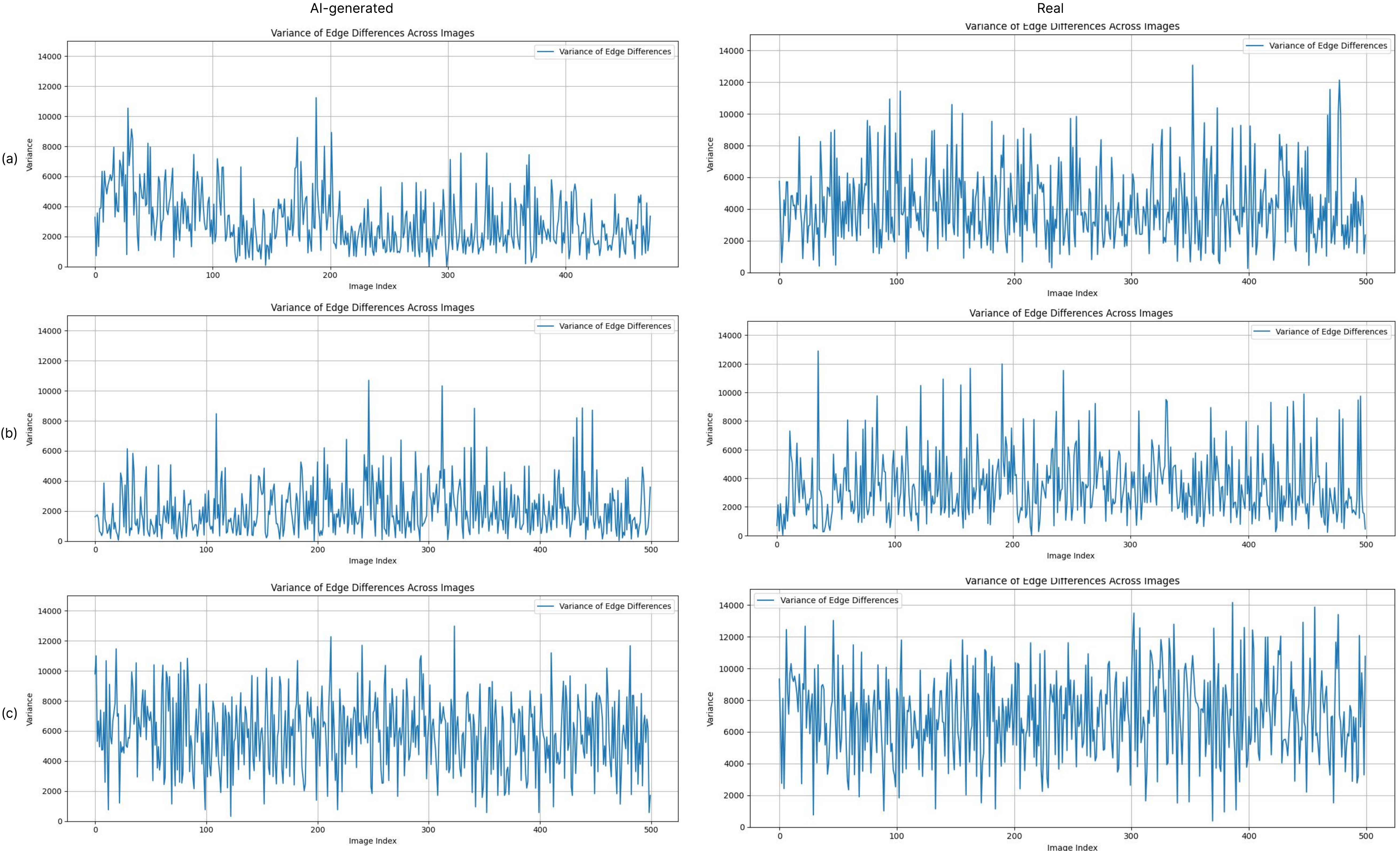}
    \caption{Variance vs. image plots on the edge difference map for (a) Custom Curated, (b) Artistic, and (c) CIFAKE datasets, with left graphs for AI-generated and right graphs for real images.}
    \label{fig:variance_map}
\end{figure*}

\begin{figure*}[htbp]
    \centering
    \includegraphics[width=1.0\linewidth]{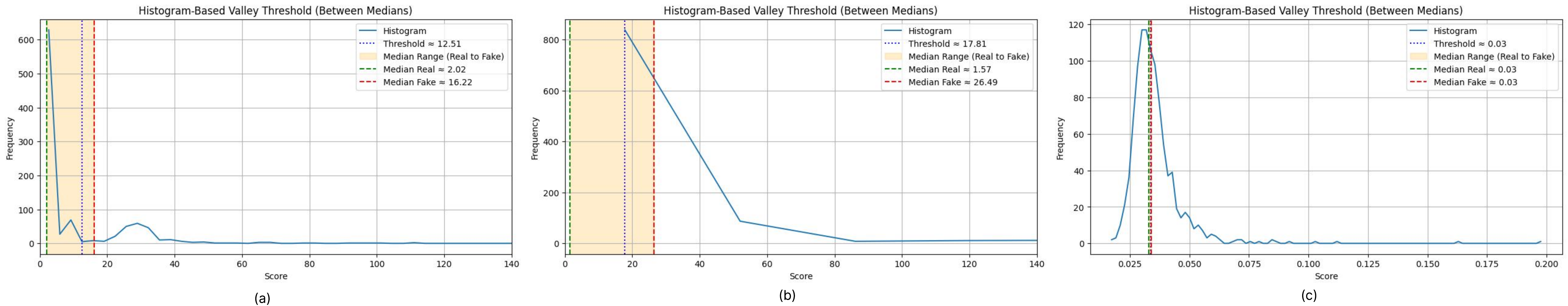}
    \caption{Valley-based threshold determination illustrating the minimum and maximum ranges, along with the selected threshold, across three datasets: (a) Custom Curated, (b) Artistic, and (c) CIFAKE.}
    \label{fig:valley_th}
\end{figure*}

Figure~\ref{fig:ed_maps} shows representative examples of input images, their denoised versions, and the corresponding edge difference maps for both real and fake samples. Figure~\ref{fig:variance_map} presents the distribution of variance values across three benchmark datasets. Finally, to support binary classification, a threshold value is estimated (Figure~\ref{fig:valley_th}) to effectively separate AI-generated from real images.

\begin{table*}[htbp]
\centering
\caption{Qualitative comparison of different architectures across multiple characteristics.}
\label{tab:qualitative_comparison}
\resizebox{\textwidth}{!}{
\begin{tabular}{lccccccc}
\hline
\textbf{Method} & \textbf{Lightweight} & \textbf{High Accuracy} & \textbf{Robustness} & \textbf{Visual Separability} & \textbf{Generalization} & \textbf{Interpretability} & \textbf{Efficiency} \\
\hline
ResNet50~\cite{resnet50}        & \ding{55} & \ding{55} & \ding{55} & \ding{55} & \ding{51} & \ding{55} & \ding{55} \\
MobileNetV2~\cite{mobilenetv2}     & \ding{51} & \ding{51} & \ding{51} & \ding{55} & \ding{55} & \ding{55} & \ding{51} \\
DenseNet121~\cite{densenet121}     & \ding{55} & \ding{51} & \ding{51} & \ding{55} & \ding{51} & \ding{55} & \ding{55} \\
EfficientNet-B0~\cite{efficientnetb0} & \ding{51} & \ding{55} & \ding{55} & \ding{55} & \ding{55} & \ding{55} & \ding{51} \\
VGG19~\cite{vgg19}           & \ding{55} & \ding{51} & \ding{55} & \ding{55} & \ding{51} & \ding{55} & \ding{55} \\
AlexNet~\cite{alexnet}         & \ding{55} & \ding{55} & \ding{55} & \ding{55} & \ding{55} & \ding{55} & \ding{55} \\
GoogLeNet~\cite{googlenet}       & \ding{51} & \ding{55} & \ding{55} & \ding{55} & \ding{55} & \ding{55} & \ding{51} \\
\textbf{ViT + EBP (Proposed)} & \ding{55} & \ding{51} & \ding{51} & \ding{51} & \ding{51} & \ding{51} & \ding{55} \\
\hline
\end{tabular}}
\end{table*}

\subsection{Ablation Study}
\subsubsection{Experimental Configurations}
\begin{figure*}[htbp]
    \centering
    \includegraphics[width=1.0\linewidth]{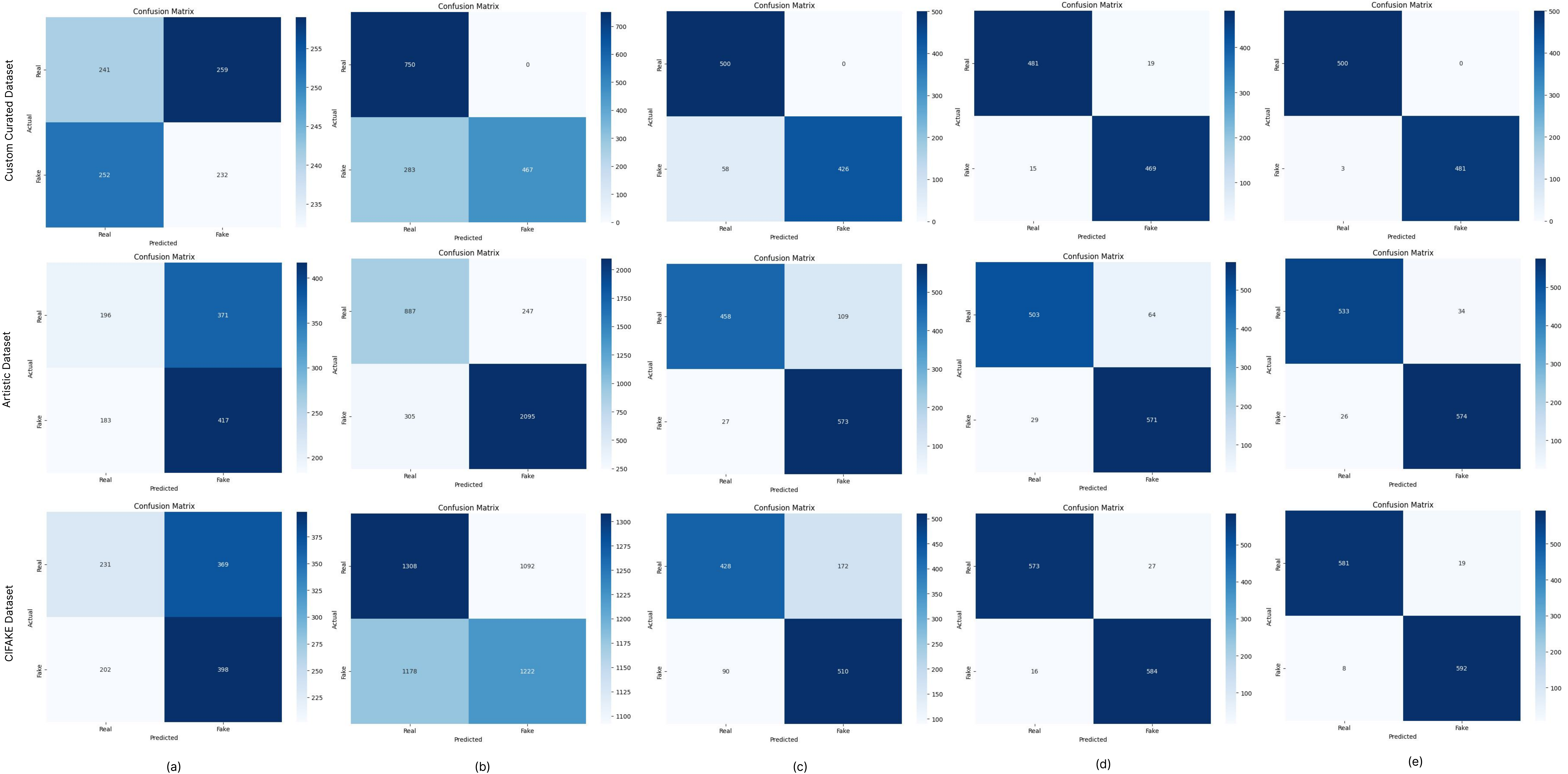}
    \caption{Confusion matrices for (a) Pretrained ViT, (b) Edge-Based Only, (c) Pretrained ViT + Post-Processing, (d) Fine-Tuned ViT, and (e) Fine-Tuned ViT + Edge-Based Post-Processing across Artistic~\cite{artistic_dataset}, CIFAKE~\cite{cifake_dataset}, and Custom datasets.}
    \label{fig:confusion_matrix}
\end{figure*}

\begin{figure*}[htbp]
    \centering
    \includegraphics[width=1.0\linewidth]{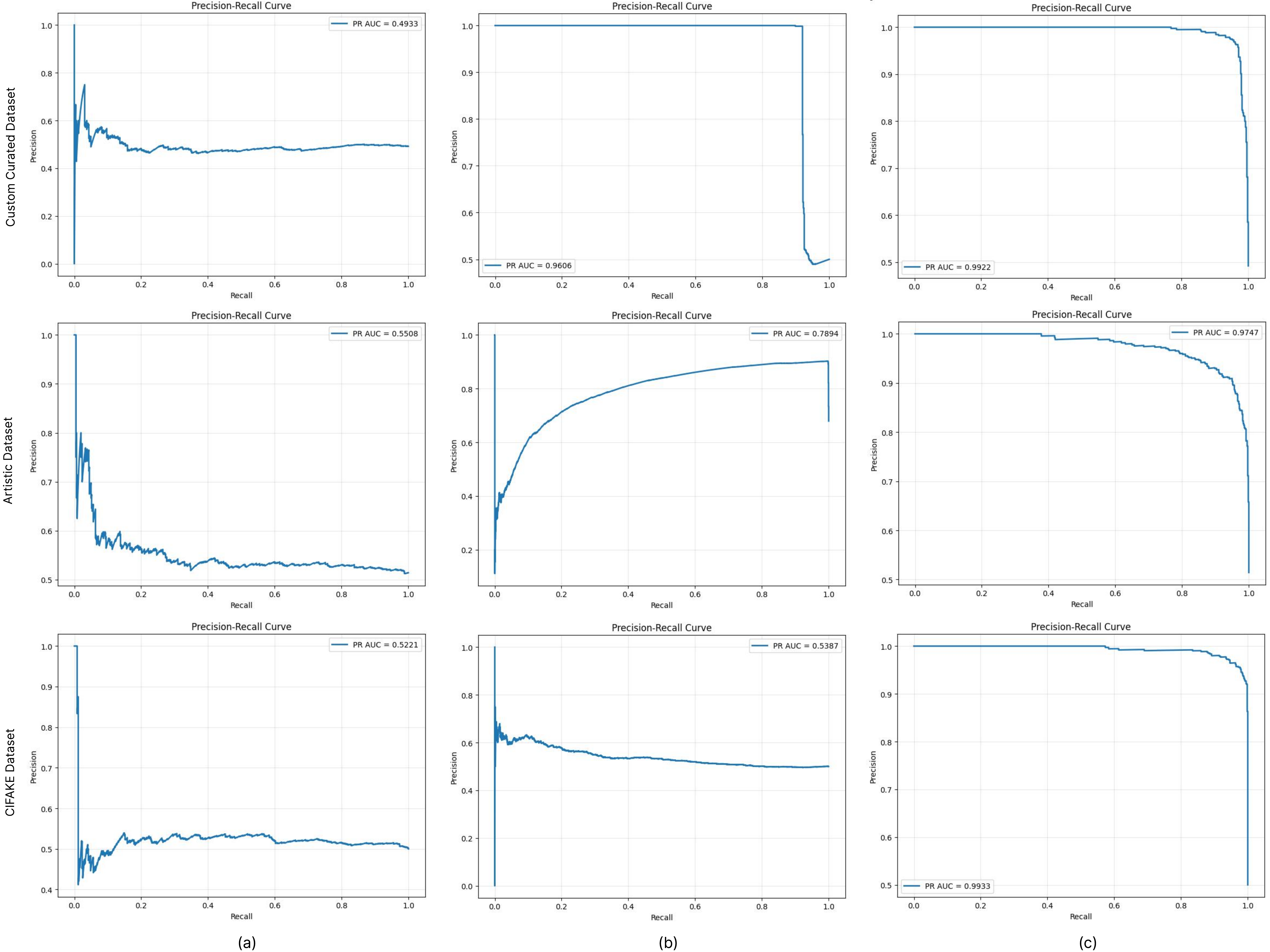}
    \caption{Precision–Recall curves for (a) Pretrained ViT, (b) Edge-Based Module, and (c) Fine-Tuned ViT across the three datasets.}
    \label{fig:pr_curve}
\end{figure*}

\begin{figure*}[htbp]
    \centering
    \includegraphics[width=1.0\linewidth]{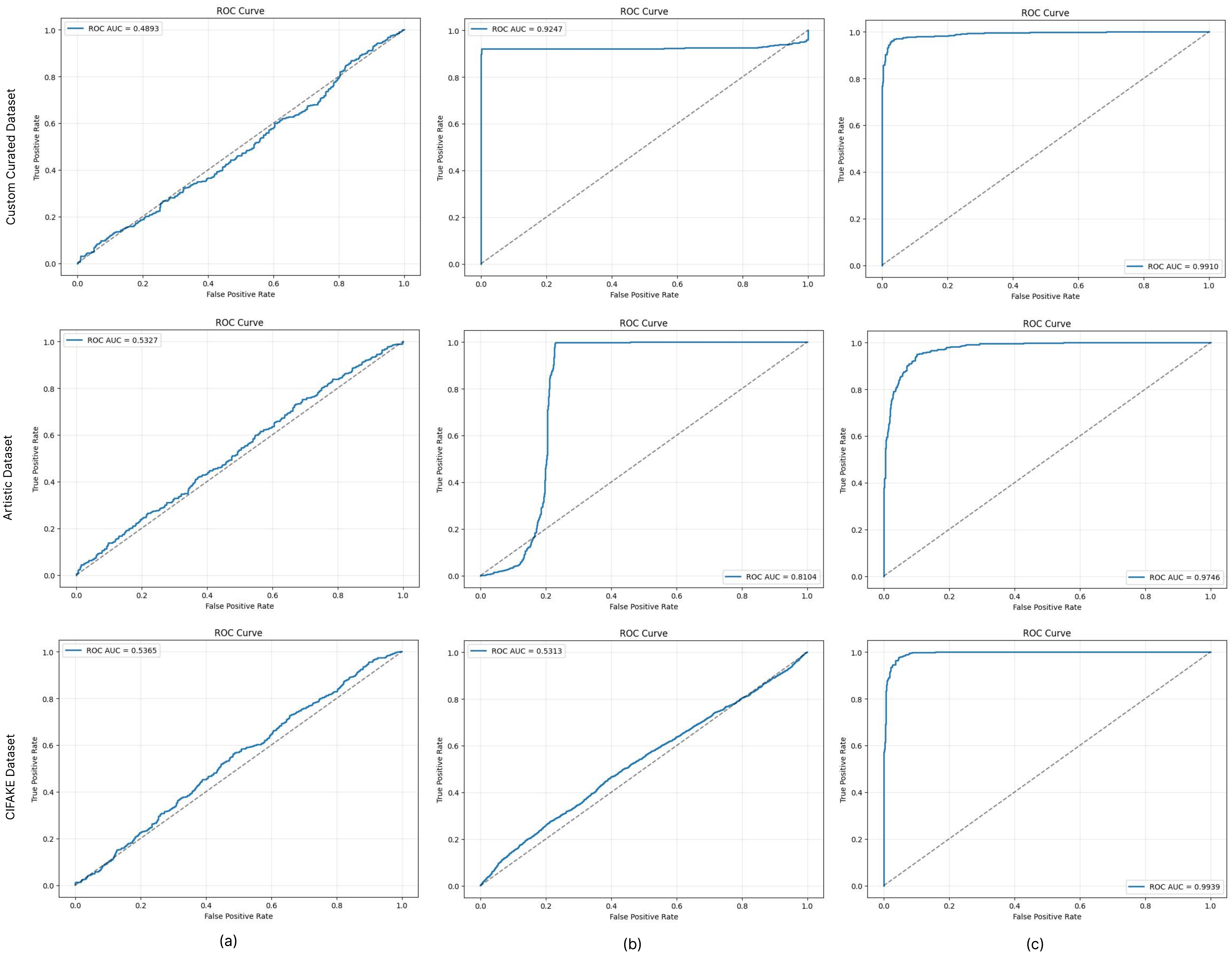}
    \caption{ROC curves for (a) Pretrained ViT, (b) Edge-Based Module, and (c) Fine-Tuned ViT across the three datasets.}
    \label{fig:roc_curve}
\end{figure*}

To evaluate the contribution of each component in our framework, we conducted an ablation study across three datasets: Artistic~\cite{artistic_dataset}, CIFAKE~\cite{cifake_dataset}, and Custom Curated. 
The analysis considers five progressively enhanced configurations, reflecting the roles of pretraining, fine-tuning, and edge-based processing. 
All models were assessed using mean IoU, precision, recall, F1-score, accuracy, and inference time. 
Quantitative results are summarized in Tables~\ref{tab:artistic_results}--\ref{tab:custom_results}, with Figures~\ref{fig:confusion_matrix}, \ref{fig:pr_curve}, and \ref{fig:roc_curve} illustrating confusion matrices, precision-recall curves, and ROC curves for each configuration.

\textbf{Pretrained ViT without Fine-Tuning.}  
This baseline employs the \texttt{vit-base-patch16-224-in21k} model~\cite{vit_base} directly on each dataset.  
Performance remains limited, with accuracies of $0.5253$, $0.5242$, and $0.4807$ on Artistic, CIFAKE, and Custom datasets, respectively.  
Mean IoU values are low (below $0.35$), demonstrating the difficulty of detecting AI-generated content without task-specific adaptation.

\textbf{Edge-Based Processing Only.}  
Classification is performed solely by the proposed edge-variance module, without ViT predictions.  
This configuration improves accuracy to $0.8438$ on Artistic and $0.8113$ on Custom dataset.  
However, performance on CIFAKE~\cite{cifake_dataset} is near chance ($0.5271$ accuracy), indicating limited cross-domain robustness.

\textbf{Pretrained ViT + Edge-Based Post-Processing.}  
Applying the edge-based post-processing module on top of pretrained ViT predictions enhances performance.  
Accuracy increases to $0.8835$ on Artistic, $0.7817$ on CIFAKE, and $0.9411$ on Custom dataset, with corresponding mean IoUs of $0.7896$, $0.6405$, and $0.8881$.  
These results highlight the effectiveness of leveraging structural edge features even without fine-tuning.

\textbf{Fine-Tuned ViT Only.}  
Fine-tuning the ViT on each dataset yields substantial improvements.  
Accuracies reach $0.9203$, $0.9642$, and $0.9654$ for Artistic, CIFAKE, and Custom datasets, respectively, with mean IoUs above $0.85$.  
This underscores the necessity of dataset-specific fine-tuning for effective feature alignment.

\textbf{Fine-Tuned ViT + Post-Processing (Proposed).}  
Combining fine-tuned ViT predictions with edge-based processing achieves the best performance across all datasets.  
Accuracy reaches $0.9486$, $0.9775$, and $0.9970$, with mean IoUs of $0.9021$, $0.9560$, and $0.9939$ for Artistic~\cite{artistic_dataset}, CIFAKE~\cite{cifake_dataset}, and Custom datasets, respectively.  
While inference time increases slightly due to edge processing, the significant gains in accuracy and IoU justify the additional cost.

Overall, the ablation study demonstrates that edge-based processing alone provides limited generalization, fine-tuning is essential for strong baseline performance, and the integration of fine-tuning with edge-aware post-processing offers consistent robustness and accuracy.

\begin{table*}[htbp]
\centering
\caption{Results of different approaches on the Artistic dataset~\cite{artistic_dataset}}
\label{tab:artistic_results}
\begin{tabular}{
  >{\raggedright\arraybackslash}p{5.0cm} 
  >{\centering\arraybackslash}p{2cm} 
  >{\centering\arraybackslash}p{1.5cm} 
  >{\centering\arraybackslash}p{1.5cm} 
  >{\centering\arraybackslash}p{1.5cm} 
  >{\centering\arraybackslash}p{1.5cm} 
  >{\centering\arraybackslash}p{1.5cm}
}
\toprule
\textbf{Approach} & \textbf{Avg. Inference Time (s)} & \textbf{Mean IoU} & \textbf{Precision} & \textbf{Recall} & \textbf{F1-Score} & \textbf{Accuracy} \\
\midrule
Pretrained ViT without Fine-Tuning & 0.0086 & 0.3454 & 0.5292 & 0.6950 & 0.6009 & 0.5253 \\
Edge-Based Processing Only         & 0.0472 & 0.7039 & 0.8945 & 0.8729 & 0.8836 & 0.8438 \\
Pretrained ViT + Post-Processing   & 0.0966 & 0.7896 & 0.8402 & 0.9550 & 0.8939 & 0.8835 \\
Fine-Tuned ViT Only                & 0.0084 & 0.8519 & 0.8992 & 0.9517 & 0.9247 & 0.9203 \\
Fine-Tuned ViT + Edge-Based Post-Processing \textbf{(Proposed)} & 0.0964 & 0.9021 & 0.9441 & 0.9567 & 0.9503 & 0.9486 \\
\bottomrule
\end{tabular}
\end{table*}

\begin{table*}[htbp]
\centering
\caption{Results of different approaches on the CIFAKE dataset~\cite{cifake_dataset}}
\label{tab:cifake_results}
\begin{tabular}{
  >{\raggedright\arraybackslash}p{5.0cm} 
  >{\centering\arraybackslash}p{2cm} 
  >{\centering\arraybackslash}p{1.5cm} 
  >{\centering\arraybackslash}p{1.5cm} 
  >{\centering\arraybackslash}p{1.5cm} 
  >{\centering\arraybackslash}p{1.5cm} 
  >{\centering\arraybackslash}p{1.5cm}
}
\toprule
\textbf{Approach} & \textbf{Avg. Inference Time (s)} & \textbf{Mean IoU} & \textbf{Precision} & \textbf{Recall} & \textbf{F1-Score} & \textbf{Accuracy} \\
\midrule
Pretrained ViT without Fine-Tuning & 0.0091 & 0.3494 & 0.5189 & 0.6633 & 0.5823 & 0.5242 \\
Edge-Based Processing Only         & 0.0111 & 0.3578 & 0.5281 & 0.5092 & 0.5185 & 0.5271 \\
Pretrained ViT + Post-Processing   & 0.0949 & 0.6405 & 0.7478 & 0.8500 & 0.7956 & 0.7817 \\
Fine-Tuned ViT Only                & 0.0081 & 0.9308 & 0.9558 & 0.9733 & 0.9645 & 0.9642 \\
Fine-Tuned ViT + Edge-Based Post-Processing \textbf{(Proposed)} & 0.0919 & 0.9560 & 0.9689 & 0.9867 & 0.9777 & 0.9775 \\
\bottomrule
\end{tabular}
\end{table*}

\begin{table*}[htbp]
\centering
\caption{Results of different approaches on the Custom Curated dataset}
\label{tab:custom_results}
\begin{tabular}{
  >{\raggedright\arraybackslash}p{5.0cm} 
  >{\centering\arraybackslash}p{2cm} 
  >{\centering\arraybackslash}p{1.5cm} 
  >{\centering\arraybackslash}p{1.5cm} 
  >{\centering\arraybackslash}p{1.5cm} 
  >{\centering\arraybackslash}p{1.5cm} 
  >{\centering\arraybackslash}p{1.5cm}
}
\toprule
\textbf{Approach} & \textbf{Avg. Inference Time (s)} & \textbf{Mean IoU} & \textbf{Precision} & \textbf{Recall} & \textbf{F1-Score} & \textbf{Accuracy} \\
\midrule
Pretrained ViT without Fine-Tuning & 0.0094 & 0.3164 & 0.4725 & 0.4793 & 0.4759 & 0.4807 \\
Edge-Based Processing Only         & 0.0256 & 0.6744 & 1.0000 & 0.6227 & 0.7675 & 0.8113 \\
Pretrained ViT + Post-Processing   & 0.0937 & 0.8881 & 1.0000 & 0.8802 & 0.9363 & 0.9411 \\
Fine-Tuned ViT Only                & 0.0080 & 0.9332 & 0.9611 & 0.9690 & 0.9650 & 0.9654 \\
Fine-Tuned ViT + Edge-Based Post-Processing \textbf{(Proposed)} & 0.0942 & 0.9939 & 1.0000 & 0.9938 & 0.9969 & 0.9970 \\
\bottomrule
\end{tabular}
\end{table*}

\subsubsection{Impact of Threshold Selection and Kernel Size}
To assess the sensitivity of our framework to key hyperparameters, we performed ablation studies on the CIFAKE dataset~\cite{cifake_dataset} focusing on two factors: (i) threshold selection strategies, and (ii) Gaussian blur kernel size. Both the edge-based only pipeline and the edge-based post-processing integrated with ViT backbones were analyzed.

\textbf{Threshold Selection.} 
We compared three strategies for threshold determination: median-based (proposed), mean-based, and mode-based values estimated via valley range analysis. As shown in Table~\ref{tab:threshold_impact}, the proposed median-based thresholding consistently outperforms the other methods in all edge-based processing configurations. For instance, in the fine-tuned ViT with edge-based post-processing, the median strategy achieves a precision of 0.9750, a recall of 0.9733, F1-score of 0.9741, and an accuracy of 97.42\%. In contrast, mean- and mode-based thresholds lead to noticeable performance drops, particularly in precision and F1-score. These results highlight the robustness of median-based thresholding in stabilizing classification performance.

\textbf{Kernel Size.} 
We further investigated the effect of Gaussian blur kernel size in the edge-based module, evaluating $3\times3$, $5\times5$, and $7\times7$ kernels. Results in Table~\ref{tab:kernel_impact} indicate that the smallest kernel ($3\times3$, proposed) yields the best overall performance across all approaches. For example, the fine-tuned ViT with edge-based post-processing achieves the highest accuracy (97.75\%) and F1-score (0.9777) with the $3\times3$ kernel. Increasing the kernel size to $5\times5$ or $7\times7$ results in reduced accuracy and recall, especially for the edge-based only setup, suggesting that excessive smoothing degrades discriminative edge information.

Together, these ablation studies confirm that careful hyperparameter selection is critical to achieving optimal results. The combination of median-based thresholding and a $3\times3$ kernel demonstrates strong robustness and provides the most reliable configuration for edge-enhanced detection with ViT backbones.

\begin{table*}[htbp]
\centering
\caption{Impact of threshold selection on the CIFAKE~\cite{cifake_dataset} dataset using valley range estimation with median, mean, and mode of image scores. Median-based thresholding (proposed) achieves the best overall performance.}
\label{tab:threshold_impact}
\begin{tabular}{lp{2.5cm}p{2.2cm}p{2.2cm}p{2.2cm}p{2.2cm}}
\toprule
\textbf{Approach} & \textbf{Metric} & \multicolumn{3}{c}{\textbf{Valley method for threshold determination}} \\ 
\cmidrule(lr){3-5}
 &  & \textbf{Th = 0.0332 (Median)} & \textbf{Th = 0.0350 (Mean)} & \textbf{Th = 0.0102 (Mode)} \\
\midrule
\multirow{4}{*}{Edge-based only} 
& Precision & 0.5281 & 0.5329 & 0.5001 \\
& Recall    & 0.5092 & 0.3983 & 1.0000 \\
& F1        & 0.5185 & 0.4559 & 0.6668 \\
& Accuracy  & 0.5271 & 0.5246 & 0.5002 \\
\midrule
\multirow{4}{*}{Pretrained ViT + Edge-based} 
& Precision & 0.7478 & 0.8009 & 0.6061 \\
& Recall    & 0.8500 & 0.8583 & 1.0000 \\
& F1        & 0.7956 & 0.8286 & 0.7547 \\
& Accuracy  & 0.7817 & 0.8225 & 0.6750 \\
\midrule
\multirow{4}{*}{Fine-tuned ViT + Edge-based \textbf{(proposed)}} 
& Precision & 0.9689 & 0.9750 & 0.9447 \\
& Recall    & 0.9867 & 0.9733 & 0.9667 \\
& F1        & 0.9777 & 0.9741 & 0.9555 \\
& Accuracy  & 0.9775 & 0.9742 & 0.9567 \\
\bottomrule
\end{tabular}
\end{table*}

\begin{table*}[htbp]
\centering
\caption{Impact of kernel size on the CIFAKE~\cite{cifake_dataset} dataset using different kernel dimensions. Smaller kernels (3×3) (proposed) consistently achieve the best overall performance across all approaches.}
\label{tab:kernel_impact}
\begin{tabular}{lp{2.5cm}p{2.2cm}p{2.2cm}p{2.2cm}p{2.2cm}}
\toprule
\textbf{Approach} & \textbf{Metric} & \multicolumn{3}{c}{\textbf{Kernel Size}} \\ 
\cmidrule(lr){3-5}
 &  & \textbf{3x3} & \textbf{5x5} & \textbf{7x7} \\
\midrule
\multirow{4}{*}{Edge-based only} 
& Precision & 0.5281 & 0.5103 & 0.4371 \\
& Recall    & 0.5092 & 0.4867 & 0.4283 \\
& F1        & 0.5185 & 0.4982 & 0.4327 \\
& Accuracy  & 0.5271 & 0.5098 & 0.4383 \\
\midrule
\multirow{4}{*}{Pretrained ViT + Edge-based} 
& Precision & 0.7478 & 0.7919 & 0.7514 \\
& Recall    & 0.8500 & 0.7800 & 0.6950 \\
& F1        & 0.7956 & 0.7859 & 0.7221 \\
& Accuracy  & 0.7817 & 0.7875 & 0.7325 \\
\midrule
\multirow{4}{*}{Fine-tuned ViT + Edge-based \textbf{(proposed)}} 
& Precision & 0.9689 & 0.9620 & 0.9652 \\
& Recall    & 0.9867 & 0.9850 & 0.9813 \\
& F1        & 0.9777 & 0.9733 & 0.9731 \\
& Accuracy  & 0.9775 & 0.9763 & 0.9752 \\
\bottomrule
\end{tabular}
\end{table*}

\subsubsection{Robustness to Degraded Inputs}
To evaluate the robustness of both the proposed model and the edge-based module, we tested images subjected to various degradations, including Gaussian noise, salt-and-pepper noise, speckle noise, Gaussian blur, motion blur, median blur, random patch occlusion, and fixed pattern occlusion. For each degradation type, a subset of real and fake images was processed, edge variance scores were computed, and thresholds were determined to measure classification accuracy under degraded conditions.

\begin{figure*}[htbp]
    \centering
    \includegraphics[width=1.0\linewidth]{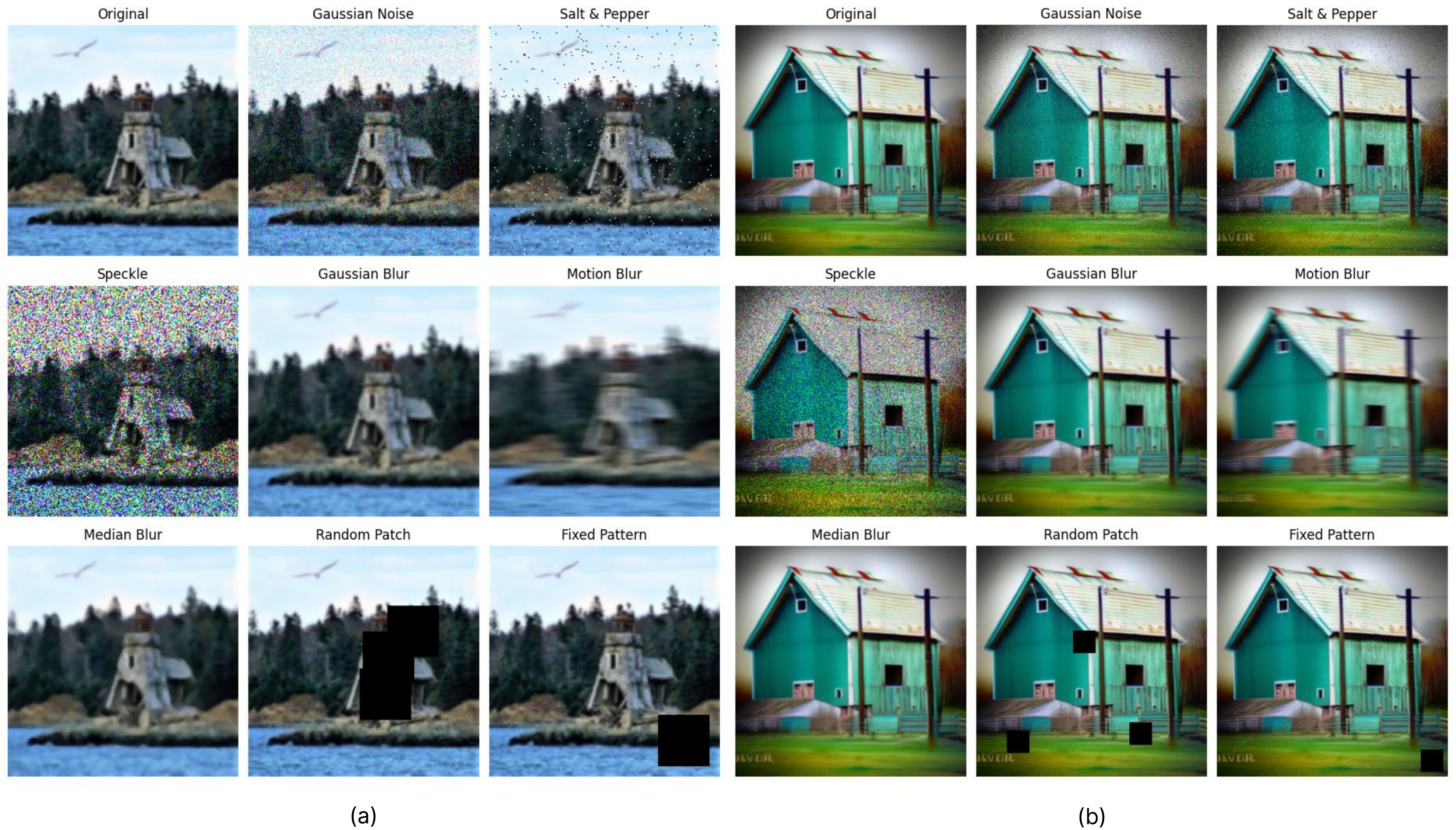}
    \caption{Examples of degraded inputs applied for robustness evaluation: (a) Real image and (b) AI-generated image under various degradation types.}
    \label{fig:degraded_examples}
\end{figure*}

As seen in Table~\ref{tab:degradation_results}, the classification accuracy varies across different degradation types, demonstrating the relative robustness of the proposed model and the edge-based module. The visual impact of these degradations is illustrated in Figure~\ref{fig:degraded_examples}.

\begin{table}[htbp]
\centering
\caption{Performance of the edge-based module within the proposed method under various image degradations.}
\label{tab:degradation_results}
\begin{tabular}{
  >{\raggedright\arraybackslash}p{2.1cm} 
  >{\centering\arraybackslash}p{1.2cm} 
  >{\centering\arraybackslash}p{1.2cm} 
  >{\centering\arraybackslash}p{0.8cm} 
  >{\centering\arraybackslash}p{1.1cm}
}
\toprule
\textbf{Degradation Type} & \textbf{Real Samples} & \textbf{Fake Samples} & \textbf{Thresh} & \textbf{Accuracy (\%)} \\
\midrule
Fixed Pattern   & 741  & 748  & 11.92 & 78.27 \\
Gaussian Blur   & 569  & 570  & 20.05 & 77.88 \\
Median Blur     & 570  & 570  & 30.22 & 74.04 \\
Gaussian Noise  & 580  & 580  & 19.06 & 74.57 \\
Motion Blur     & 570  & 570  & 8.46  & 80.79 \\
Random Patch    & 770  & 779  & 13.98 & 76.44 \\
Salt \& Pepper  & 1157 & 1187 & 14.20 & 75.85 \\
Speckle Noise   & 568  & 570  & 8.62  & 75.92 \\
\bottomrule
\end{tabular}
\end{table}


\section{Discussion}
The experimental results validate the effectiveness of the proposed hybrid framework in detecting AI-generated images. As summarized in Table~\ref{tab:sota_comparison}, the integration of a fine-tuned Vision Transformer (ViT) with the edge-based processing (EBP) module significantly outperforms existing state-of-the-art architectures in terms of classification accuracy and reliability. The proposed method achieves an accuracy of 97.75\% and an F1-score of 97.77\%, surpassing widely adopted models such as DenseNet121~\cite{densenet121} (93.87\%), VGG19~\cite{vgg19} (93.75\%), and EfficientNet-B0~\cite{efficientnetb0} (92.75\%). These improvements highlight the complementary strengths of the hybrid framework: ViT captures domain-specific high-level semantic features, while the edge-based module leverages structural cues through variance analysis of edge-difference maps.

Another notable contribution of the edge-based module is its interpretability and computational efficiency. Unlike deep learning-only approaches that rely heavily on large-scale parameter tuning, the EBP operates independently of training, requiring minimal computation. Even when used in isolation, the EBP demonstrates competitive performance, underscoring its potential as a lightweight and standalone detection tool. However, its integration with ViT further amplifies detection robustness, particularly by addressing cases where either approach alone may fail.

Despite its strong performance, the edge-based module has limitations. When applied to highly complex images with sharp artificial details, especially those generated by advanced vision transformer-based generative models, the edge variance scores between real and synthetic images become less distinguishable. Similarly, in scenarios where AI-generated images closely mimic real images at pixel-level sensitivity, the edge-based module alone cannot provide reliable classification. These challenges emphasize that while the EBP adds interpretability and lightweight detection capabilities, it is best utilized in combination with ViT to balance sensitivity and generalization. Overall, the hybrid framework demonstrates superior robustness compared to existing models, reinforcing the value of integrating fundamentally different detection strategies.

\section{Conclusion}
\label{sec:conclusion}
This work introduced a hybrid framework for AI-generated image detection that combines a fine-tuned Vision Transformer with a novel edge-based processing module. The edge-based module computes variance from edge-difference maps before and after denoising, enabling interpretable and computationally lightweight discrimination between real and synthetic content. When integrated with ViT predictions, the framework effectively captures both high-level semantic features and fine-grained structural inconsistencies, achieving state-of-the-art performance across CIFAKE~\cite{cifake_dataset}, Artistic~\cite{artistic_dataset}, and Custom Curated datasets.

Quantitative evaluations demonstrate that the proposed method consistently outperforms conventional CNN-based architectures, achieving the highest accuracy and F1-score among compared models (Table~\ref{tab:sota_comparison}). While the edge-based module alone offers efficiency and interpretability, its limitations in handling complex or highly realistic transformer-generated images highlight the necessity of hybrid integration. By unifying complementary perspectives, the proposed framework provides a practical, scalable, and accurate solution for digital forensics and content authentication. Its lightweight design and adaptability to both static images and video frames make it particularly suitable for real-world deployment in combating the proliferation of AI-generated content.




\section*{Declaration of Funding}
This paper was not funded. 


\section*{Author Contributions}
\textbf{Dabbrata Das:} Conceptualization, Methodology, Visualization, Writing – original draft, and Project administration. \textbf{Mahshar Yahan:} Methodology, Formal analysis, Writing – original draft \& review. \textbf{Md Tarek Zaman:} Formal analysis, Dataset preparation, and Review. \textbf{Md Rishadul Bayesh:} Investigation, Dataset preparation, and Review.

\section*{Ethical Approval}
Not required.

\section*{Declaration of Competing Interest}
The authors declare no competing interests.

\section*{Acknowledgements}
This work is supported in part by the Uttara University.



\bibliographystyle{cas-model2-names}

\bibliography{refs}

\end{document}